\documentclass[final]{cvpr}

\usepackage{times}
\usepackage{epsfig}
\usepackage{graphicx}
\usepackage{amsmath}
\usepackage{amssymb}
\usepackage{multirow}
\usepackage{animate}
\usepackage{subfigure}
\usepackage{multirow}
\usepackage{booktabs}

\usepackage[pagebackref=true,breaklinks=true,colorlinks,bookmarks=false]{hyperref}

\begin{document}

\title{CRFormer: A Cross-Region Transformer for Shadow Removal}
\author{Jin Wan$^{1}$, Hui Yin$^{1,*}$, Zhenyao Wu$^{2}$, Xinyi Wu$^{2}$, Zhihao Liu$^{3}$, Song Wang$^{2,}$\thanks{Co-corresponding authors.}\\{\normalsize $^1$Beijing Jiaotong University \qquad $^2$University of South Carolina\qquad $^3$ China Mobile Research Institute}\\{\tt\small\{jinwan,hyin\}@bjtu.edu.cn,\{xinyiw,zhenyao\}@email.sc.edu}\\{\tt\small\ liuzhihao@chinamobile.com,songwang@cec.sc.edu}
}

\maketitle
\pagestyle{empty}
\thispagestyle{empty}

\begin{abstract}
Aiming to restore the original intensity of shadow regions in an image and make them compatible with the remaining non-shadow regions without a trace, shadow removal is a very challenging problem that benefits many downstream image/video-related tasks. 
Recently, transformers have shown their strong capability in various applications by capturing global pixel interactions and this capability is highly desirable in shadow removal.
However, applying transformers to promote shadow removal is non-trivial for the following two reasons: 1) The patchify operation is not suitable for shadow removal due to irregular shadow shapes; 
2) shadow removal only needs one-way interaction from the non-shadow region to the shadow region instead of the common two-way interactions among all pixels in the image.
In this paper, we propose a novel cross-region transformer, namely CRFormer, for shadow removal which differs from existing transformers by only considering the pixel interactions from the non-shadow region to the shadow region without splitting images into patches. This is achieved by a carefully designed region-aware cross-attention operation that can aggregate the recovered shadow region features conditioned on the non-shadow region features. Extensive experiments on ISTD, AISTD, SRD, and Video Shadow Removal datasets demonstrate the superiority of our method compared to other state-of-the-art methods. 
\end{abstract}


\section{Introduction}
With the growing use of various cameras, the digital images/videos are everywhere to keep records of faces, documents, wonderful moments where undesirable shadows may show up and degrade the visual quality \cite{Bako16documentS,He_21mmshadowFace,Lin2020documentS,liu2021from,Fu_2021_Auto}.
Shadows also impact the feature representation of images and may unfavorably affect subsequent image/video processing tasks such as object detection and tracking~\cite{Girshick_2014_rcnn,Girshick2015FaRcnn}, intrinsic image decomposition~\cite{Li2018LearningII,Liu2020UnsupervisedLF}, and semantic segmentation~\cite{Shelhamer2017FCN,Caelles2017OneShotVO,Wu_2021_CVPR}. 
To both improve the image quality and benefit the downstream tasks, shadow removal is highly desirable, whose goal is to recover the pixel intensity of shadow regions cast by objects.  
It is a challenging problem due to complex lighting conditions and irregular shadow shapes.

Owing to the advancement of the deep convolutional neural networks (CNNs) and the extracted representative deep features, CNN-based methods ~\cite{qu2017deshadownet,wang2018stacked,hu2019direction,Le2019Shadow,le2020from,Fu_2021_Auto,liu2021from} become the mainstream for shadow removal by exhibiting superior performance over traditional methods~\cite{shor2008shadow,guo2012paired,Shechtman2016Appearance}.
As mentioned in ~\cite{zhu2018bidirectional,Chen_2021_CANet}, image contextual cues from non-shadow regions are crucial for shadow removal.
Unfortunately, {most of the existing CNN-based approaches are ineffective in modelling long-range pixel dependencies for the large receptive fields due to the convolution operation. }
Therefore, the information from the non-shadow regions is not fully explored to recover each pixel of shadow regions in these methods. 
Recently, {a context-aware CNN-based approach}~\cite{Chen_2021_CANet} attempts to perform shadow removal by matching feature similarity between shadow and non-shadow patches and then transferring the contextual information of paired patches to help relieve this issue.
However, it requires an extra sizeable patch-based dataset to train the contextual patch matching module as a prerequisite, which is time-consuming and laborious. And from the methodology perspective, it only picks top-3 {(in the original publication)} similar patches from the non-shadow region for contextual information transfer, which is still not conducive to exploiting all the pixel-level information from the non-shadow region for shadow removal.

Recently, Transformer has been successful in many computer vision tasks~\cite{dosovitskiy2021an,guo2021image,li2021pose,liuswinT,pu2022edter,botach2021end}, where long-range contextual information can be effectively modeled. We thus consider taking the advantage of the transformer to enhance the connection from the non-shadow region to the shadow region.
However, there are two main challenges to be addressed before using transformers for shadow removal. Firstly, transformers are mainly proposed to take image patches as input, which is not intuitively suitable for shadow removal due to the irregular shadow shapes cast by objects.
Secondly, the global pixel interactions of existing transformers consider the contributions of all pixels for recovering the shadow region, only the one-way connection from the non-shadow region to the shadow region should be considered for shadow removal because of the corrupted features from the shadow region.

\begin{figure*}[htbp]\small
	\centering
	\includegraphics[width=1.0\linewidth]{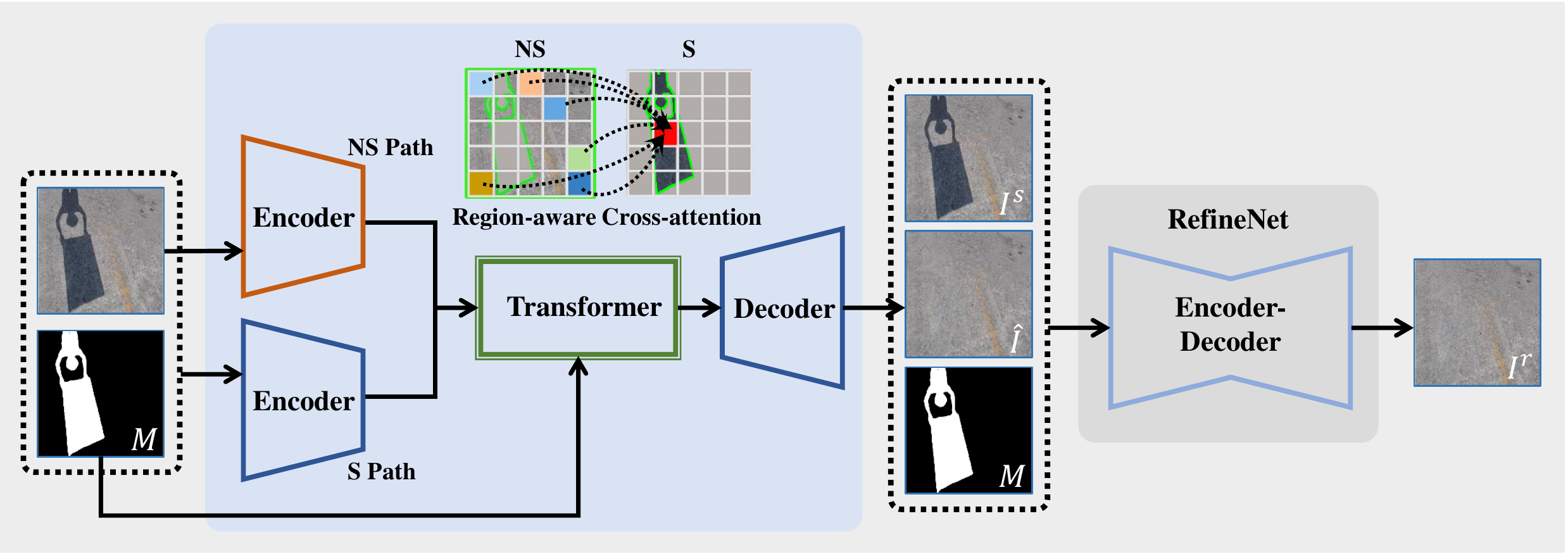}
	\vspace{-.3cm}
	\caption{
	Overview of the proposed CRFormer framework-- NS: non-shadow; S: shadow. The dual encoder takes the shadow image $I^{s}$ and its shadow mask $M$ as input with each encoder processing different input, \textit{i.e.,} the input of the top one is the three-channel $I^{s}$ and the bottom one is fed with the four-channel concatenation of $I^{s}$ and $M$.
    Then, a Transformer layer~(see Fig.~\ref{fig:2}) with our proposed region-aware cross-attention is utilized to aggregate the restored shadow region features conditioned on the non-shadow region features, and the output is fed into a single decoder to obtain the de-shadowed result $\hat{I}$.
    Finally, $\hat{I}$ is composed with $I^{s}$ and $M$ sent into the RefineNet to generate the refined shadow-removal result $I^{r}$.
	}
	\label{fig:1}
	\vspace{-0.3cm}
\end{figure*}

To address the above issues, we propose a novel cross-region transformer (CRFormer) for shadow removal in the form of a hybrid CNN-transformer framework (see Fig.~\ref{fig:1}).
A CNN-based dual encoder is firstly employed to extract asymmetrical features between the two paths given the shadow image and its shadow mask.
{Then, the proposed Transformer layer with $N$ cross-region alignment blocks takes in the features from both shadow and non-shadow regions to build connections from the non-shadow regions to the shadow regions, which is achieved by a newly designed region-aware cross-attention. In this way, the proposed CRFormer can utilize adequate contextual information from non-shadow regions to recover the intensity of each shadowed pixel in shadow regions.} 
After that, the output of a series of cross-region alignment blocks is fed into a single CNN-based decoder to achieve the de-shadowed result. Finally, we utilize a light U-shaped network for post-processing to refine the obtained shadow-removal results. 
We evaluate our proposed CRFormer on the ISTD, AISTD, SRD, and Video Shadow Removal datasets. Extensive experiments demonstrate the effectiveness and superiority of the proposed method.

The main contributions of our paper are as follows:
\begin{itemize}
\item We propose a novel hybrid CNN-transformer framework, named cross-region transformer (CRFormer), for high-quality shadow removal.
In CRFormer, all the pixels from the non-shadow regions are taken into account to help recover each shadowed pixel, which fully exploits the underlying contextual cues from non-shadow regions for shadow removal.
To the best of our knowledge, it is the first transformer-based framework tailored for shadow removal. 
\item We propose a novel region-aware cross-attention (RCA) which aggregates the pixel-level features of non-shadow regions into the recovered shadow region features by only building a one-way connection from the non-shadow region to the shadow region.

\item Extensive experiments demonstrate the superiority of the proposed method over the state-of-the-art methods on four well-known datasets, including ISTD, AISTD, SRD, and Video Shadow Removal. 
\end{itemize}

\section{Related Work}
\label{sec:related-work}

\subsection{Shadow Removal}
\label{subsec:sr}
Early studies in shadow removal usually focus on using low-level image representations to remove shadows, including gradient statistics~\cite{gryka2015learning} and illumination prior~\cite{shor2008shadow,guo2012paired,zhang2015shadow,Shechtman2016Appearance}, \textit{etc}. 
After that, convolutional neural networks (CNNs) based methods become the mainstream for shadow removal ~\cite{qu2017deshadownet,wang2018stacked,hu2019direction,Le2019Shadow,le2020from,Le_tpami21,hu2019mask,Cun_Pun_Shi_2020_DHAN,Fu_2021_Auto,liu2021from} and achieve significantly better performance than the traditional methods. 
For example, DeshadowNet~\cite{qu2017deshadownet} exploits multi-context information and predicts shadow matte layer to remove shadows. ST-CGAN~\cite{wang2018stacked} and DSC~\cite{hu2019direction} combine shadow detection and removal to boost the performance. SP+M-net~\cite{Le2019Shadow}, Param+M+D-Net~\cite{le2020from}, and SP+M+I-net~\cite{Le_tpami21} decompose shadow images into different learnable parameters to 
achieve shadow-free images, which roughly exploits the global (both shadow and non-shadow regions) context information for shadow removal without fully exploring the underlying relationship between shadow and non-shadow regions. Mask-shadowGAN~\cite{hu2019mask}, G2R-ShadowNet~\cite{liu2021from}, and DC-ShadowNet~\cite{jin2021dc} utilize the GAN model to perform shadow removal from the perspective of image generation by learning the mapping between shadow and shadow-free domains. Recently, DHAN~\cite{Cun_Pun_Shi_2020_DHAN} removes boundary artifacts of de-shadowed images by using dilated convolutions to enlarge the receptive field to aggregate dilated multi-context features. AEFNet~\cite{Fu_2021_Auto} generates shadow-removal images by adaptively fusing the estimated multiple over-exposed images. 

However, the above CNN-based methods do not explicitly make use of the contextual information from non-shadow regions to remove the shadows in the shadow image. Besides, the convolution operation is insufficient for modelling long-range pixel dependencies, which limits the extraction of global information for shadow removal.
To solve this problem, CANet~\cite{Chen_2021_CANet} transfers the contextual information of non-shadow regions to the shadow region in a patch-based fashion through the contextual patch matching and contextual feature transfer module, where unpaired data in form of patches are cropped from the same shadow image under the shadow mask. However, this patch-based method suffers from a heavy computational load because of the repetitive cropping of a small step size as mentioned in~\cite{liu2021from}. In addition, this method only uses the top-3 (in their original paper) similar non-shadow region patches to transfer the contextual information into the matching shadow region patches, which does not fully leverage all the pixel-level information from non-shadow regions.

Different from all the above methods, our method is a hybrid framework of the CNN and transformer and we take all non-shadowed pixels into account to recover each shadowed pixel.


\subsection{Vision Transformer}
\label{subsec:transformer}
Transformer, a neural network mainly based on self-attention mechanism, was firstly proposed for natural language processing tasks~\cite{vaswani2017attention,Devlin2019BERTPO}. It has recently been extended to and shown success in multiple image/video-related tasks due to its powerful long-range dependency modelling capabilities. ViT~\cite{dosovitskiy2021an} proposes to break the image into patches and directly feed these patches (tokens) into a pure transformer for image classification. Currently, its variants are also widely used in other applications~\cite{li2021diverse,liuswinT,zheng2021rethinking,pu2022edter}. 
More recently, DETR~\cite{carion2020end} utilizes transformer in combination with CNN and shows excellent performance in object detection. This architecture brings direct inspiration to other recent works ~\cite{dai2021up,kim2021hotr,li2021pose,zou2021end,zhang2021few,guo2021image,li2021revisiting}. 

Our study is inspired by the aforementioned pioneer works but differs from them significantly in two respects. First, we make the first attempt to extend the transformer into shadow removal by taking the property of shadows into account. Second, our key idea is to only perform one-way region-specific pixel connections from one region to another, which is achieved by region-aware cross-attention instead of global self-attention employed by existing transformers, resulting in favorable de-shadowed results.

\section{Shadow Removal with Transformer}
\label{sec:method}
\subsection{Overview}
\label{subsec:overview}
The overall pipeline of the proposed cross-region transformer (CRFormer) is shown in Fig.~\ref{fig:1}, which is a hybrid CNN-transformer framework. 
With the input shadow image $I^{s}$ and its shadow mask $M$, a dual-encoder architecture is designed in CRFormer to extract asymmetrical features which involve different region features of interest.
The input of the non-shadow path is the three-channel input image {that contains contextual information from non-shadow regions} and the shadow path is the four-channel concatenation of the input image and its shadow mask. Note that the network design of the two encoders is different.
To reduce the interference between the shadowed and non-shadowed pixels due to deeper convolution, \textit{i.e.,} extracting pure features inside each region to accurately provide non-shadow region features of interest, the top encoder (non-shadow path) is built on a shallow sub-net using only three convolutions, which includes two $3\times3$ average-pool convolutions to downsample the feature maps and one $1 \times 1$ convolution to adjust the dimension of the feature maps to match that of bottom encoder output.
The bottom encoder of the shadow path instead is a deeper one that consists of several convolutions and residual blocks where two convolutions with stride set to 2 to downsample the feature maps. The effectiveness of the dual encoder is verified in Table~\ref{tab:abl_loss}.


Given that both shadow and non-shadow regions of an image are usually parts of the same harmonious scene, it is expected that the contextual information of non-shadow regions can help recover the shadow-free pixel intensity of shadow regions. To achieve this goal, we design a new Transformer layer with region-aware cross-attention as detailed in Section~\ref{subsec:Region-transformer}.
Afterwards, its output is sent to a decoder to reconstruct the de-shadowed image $\hat{I}$.
At last, we perform a result refinement to improve the quality of the de-shadowed result. This is achieved by first obtaining a composite image 
$I^ {c} = M \circ \hat{I} + (1 - M) \circ I^ { s}$, where $\circ$ is the Hadamard product. 
By taking the $I^ { c}$ and $M$ as input, the RefineNet is executed to generate the final shadow-removed image $I^ { r}$, which is a similar post-processing step as in~\cite{liu2021from,Le_tpami21,Chen_2021_CANet}. 
We use the U-shape network in~\cite{liu2021from} as the backbone of RefineNet and remove its half of the filters to reduce computational complexity.  

\subsection{Review of Transformer}
\label{subsec:Revisiting-transformer}
The initial transformer~\cite{vaswani2017attention} consists of $N$ encoder blocks. Each block consists of multi-head attention (MHA), multi-layer perceptron (MLP) and layer normalization (LN). Additionally, a residual connection is utilized at the end of each block to prevent network degradation. Generally, MHA executes multiple attention modules in parallel and projects the concatenated output. Most importantly, in each attention module, attention map is calculated by applying the dot product similarity to a set of Query vectors and Key vectors, and using the results to re-calibrate a set of Value vectors to achieve the aggregated outputs. All these operations can be formulated as:
\begin{equation}\label{eq:1}
\centering
\begin{split}
F_a &= softmax( \frac{Q{K^T}}{\sqrt{d}}) V, \\
Q &= F_q W_q, \qquad K = F_{kv} W_k, \qquad V = F_{kv} W_v,
\end{split}
\end{equation}
where $F_q$ and $F_{kv}$ represent feature descriptors of Query and Key/Va-lue, respectively, $ W_q, W_k, W_v \in \mathbb{R}^{C \times d}$ are linear learnable matrices, $C$ is the dimension of token embeddings, and
$d$ is the dimension of $Q$, $K$, and $V$. When $F_q=F_{kv}$, it is equivalent to the self-attention mechanism~\cite{vaswani2017attention}. 

\begin{figure*}[htbp]\small
	\centering
	\includegraphics[width=1.\linewidth]{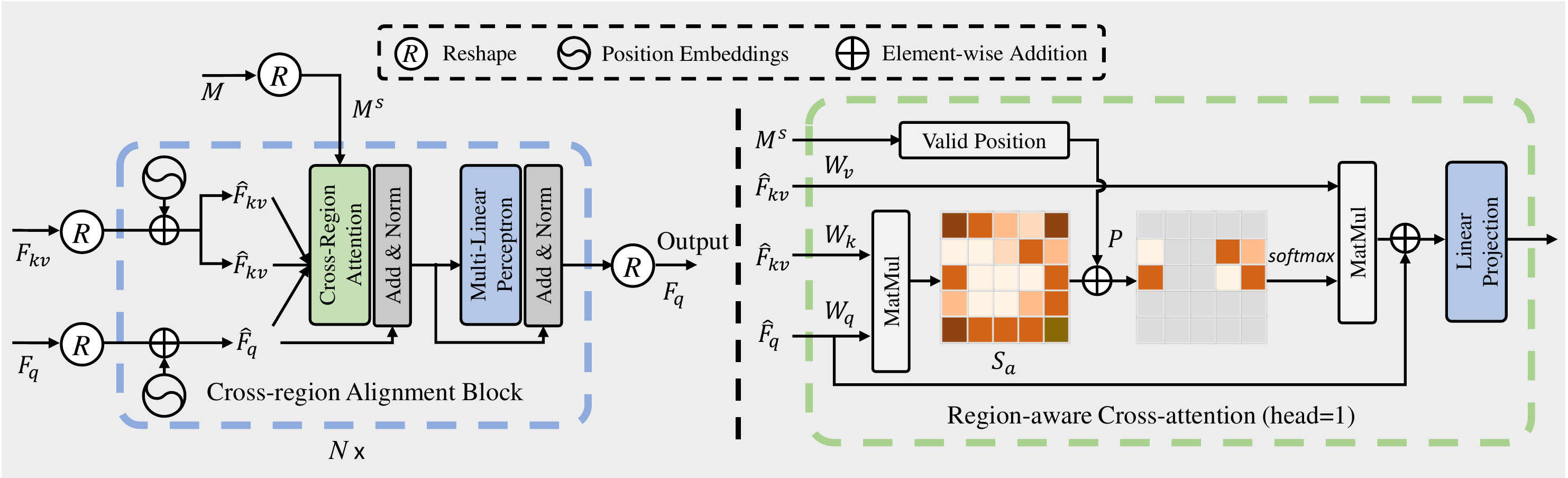}
	\vspace{-0.3cm}
	\caption{
	An illustration of the proposed Transformer layer with the detail computation pipeline of region-aware cross-attention. The Transformer layer consists of $N$ cross-region alignment blocks to transfer contextual information from non-shadow into shadow regions. And in each cross-region alignment block, the region-aware cross-attention is to compute the one-way attention from non-shadowed pixels to shadowed pixels. 
	}
	\label{fig:2}
	\vspace{-0.3cm}
\end{figure*}

\subsection{Transformer with Region-aware Cross-attention}
\label{subsec:Region-transformer}
To recover the shadowed pixels, it is critical to fully explore and exploit the underlying contextual cues of non-shadow regions. We thus propose a novel Transformer layer with region-aware cross-attention (RAC) to transfer adequate contextual information from non-shadow to shadow regions. Inside the Transformer layer, there are $N$ cross-region alignment blocks.
The detailed pipeline of the cross-region alignment block and the RAC computation are shown in Fig.~\ref{fig:2}.  
Specifically, given the compact shadow features $F_q \in \mathbb{R}^{H \times W \times C}$ and non-shadow features $F_{kv} \in \mathbb{R}^{H \times W \times C} $ of interest produced by the CNN-based dual encoder architecture, we first flatten them into $1D$ vectors (with a shape of $HW \times C$) as the input of the cross-region alignment block, where $H$, $W$, and $C$ are the height, width, and channel of the feature maps, and we design here that the feature shapes of $F_q$ and $F_{kv}$ are equal. 
In addition, we take pixels (corresponding to image patches in~\cite{dosovitskiy2021an}) as the input tokens of the Transformer layer, and the channel of each pixel as token embedding. We then perform positional encoding for each token, which can be represented as
\begin{equation}\label{eq:2}
\centering
\begin{split}
\hat{F}_{q} = F_{q} + P_{q}, \qquad \hat{F}_{kv} = F_{kv} + P_{kv},
\end{split}
\end{equation}
where $P_q, P_{kv} \in \mathbb{R}^{HW \times C}$ are the positional embeddings as in~\cite{carion2020end}.

Afterwards, we employ region-aware cross-attention (see Section~\ref{subsubsec:Region-attention}) to perform the one-way attention from non-shadowed pixels to shadowed pixels for aggregating relevant non-shadow region features into restored shadow region features. Note that we do not integrate the global features (\textit{i.e}, both shadow and non-shadow region features) to the de-shadowed region features through the cross-region alignment block, because shadow region features might be corrupted by shadows and bring a negative impact to the final de-shadowed feature aggregation as proved in Table~\ref{tab:abl_att}.
Finally, the aggregated features by the region-aware cross-attention go through the MLP and the LN layers to produce an output of the cross-region alignment block. When stacking $N$ blocks, the output of the preceding block is fed into the next one.

\subsubsection{Region-aware Cross-attention}
\label{subsubsec:Region-attention}

According to the above discussion, to aggregate the pixel-level features of non-shadow regions into recovered features of shadow regions and avoid confusion of irrelevant image features (\textit{i.e}., features cracked by shadows), we propose a region-aware cross-attention operation to conduct one-way pixel interactions from non-shadow to shadow regions. The detailed computation steps are illustrated in Fig.~\ref{fig:2} (right).
The attention is computed using $\hat{F}_q$, $\hat{F}_{kv}$, and the resized and shaped shadow mask $M^s \in \mathbb{R}^{HW \times 1}$. Note that due to the heavy computational complexity of the multi-head attention operation, we set the number of heads to $1$ to perform region-aware cross-attention in this paper.

Formally, the Query vector, Key vector, and Value vector are initially obtained by multiplying $\hat{F}_q$ and $\hat{F}_{kv}$ with three linear learnable matrices ($W_q, W_k,$ and $W_v$) as in Eq.~(\ref{eq:1}).
We then compute the similarity score map $S_a \in \mathbb{R}^{HW \times HW}$ to measure the correlation between all Query and Key pixels/tokens. { After that, to better explore non-shadow region features for recovering the shadowed pixels, we only focus on the query pixels located at $i~(i \in \left\{ 0,1,...,(HW-1) \right\})$ in the shadow region and the key pixels at $j~(j \in \left\{ 0,1,...,(HW-1) \right\})$ in the non-shadow region. Given the reshaped binary shadow mask $M^s$, the region-aware cross-attention operation is applied from \textit{j} to \textit{i} if ${M^s}_{(i)}~=~1~and~{{M^s}}_{(j)}~= 0$. Thus, by traversing all tokens of Query and Key, we can obtain a map $P \in \mathbb{R}^{HW \times HW}$ to indicate the positional correspondence between non-shadowed pixels and shadowed pixels in $S_a$,}
which is defined as an additive bias
\begin{equation}\label{eq:4}
\centering
\begin{split}
{P}_{(i, j)} = \begin{cases}
	0, & if~{M^s}_{(i)}=1~and~{M^s}_{(j)}=0 , \\
	-\infty, & otherwise .
	\end{cases}
\end{split}
\end{equation}

After that, for all query pixels/tokens in shadow regions, the aggregated features re-calibrated by all key pixels/tokens from non-shadow regions are computed by
\begin{equation}\label{eq:5}
\centering
\begin{split}
F_{rca} &= RCAtten(Q, K, V, M^s) \\
    &= softmax(S_{a} + P )V \\ 
    &= softmax( \frac{Q {K^T}}{\sqrt{d}} + P ) V. 
\end{split}
\end{equation}
Finally, $\hat{F}_q$ is added to $F_{rca}$ for preventing the degradation of the non-shadow region information and then sent to a linear projection layer for forwarding propagation to obtain the final output. In this manner, the Transformer layer can effectively avoid the attention being biased by irrelevant features destroyed by shadows and transfer appropriate contextual information from non-shadow to shadow regions for reconstructing high-quality de-shadowed results.

\subsection{Loss Function}
\label{subsec:loss}
The proposed CRFormer is trained in an end-to-end fashion.
The total loss consists of the reconstruction loss $\mathcal{L}_{rec}$ and the spatial loss $\mathcal{L}_{spa}$, which is defined as
\begin{equation}
\mathcal{L}=\omega_1 \mathcal{L}_{rec} + \omega_2 \mathcal{L}_{spa},
\end{equation}
where $\omega_1$ and $\omega_2$ are the weights of different loss items and we empirically set them to 1 and 10 in our experiments, respectively. Specifically, the pixel-wise $L1$ distance is employed to ensure that the pixel intensities of our de-shadowed results $\hat{I}$ and $I^{r}$ are consistent with the corresponding ground-truth image $I^{gt}$, which is calculated as
\begin{equation}
\mathcal{L}_{rec}=\Vert \hat{I} - I^{gt}\Vert_1 + \Vert  I^{r} - I^{gt}\Vert_1.
\end{equation}
In addition, inspired by low-light image enhancement task~\cite{guo2020zero}, we introduce the spatial consistency loss to enforce the spatial consistency of the image by preserving the differences between adjacent regions of the de-shadowed image and its corresponding shadow-free version,  which is expressed as
\begin{equation}
\begin{split}
\mathcal{L}_{ spa} &= \phi(\hat{I}, I^{gt}) + \phi(I^{r}, I^{gt}), \\
\phi(\cdot , \cdot)&=\frac{1}{L}\sum^L_{x=1}\sum_{y\in \Omega(x)}(|(A_x, A_y)|-|(B_x, B_y)|)^2,
\end{split}
\end{equation}
where $L$ represents the number of local areas, $\Omega(x)$ indicates four areas (up, down, left, and right) near the local area $x$, and  $A$ and $B$ are the average values of the local areas of de-shadowed image and its corresponding ground-truth image, respectively.

\section{Experiments}
\label{sec:exps}

\subsection{Datasets}
\label{subsec:dataset}
We conduct experiments on four popular datasets: ISTD~\cite{wang2018stacked}, AISTD \cite{Le2019Shadow}, SRD~\cite{qu2017deshadownet}, and Video Shadow Removal~\cite{le2019weakly}. 

\textbf{ISTD}~\cite{wang2018stacked} and \textbf{AISTD}~\cite{Le2019Shadow} include 1,870 natural shadow images, 1,330 for training and 540 for testing. Each shadow image has the corresponding shadow-free image and shadow mask image. 
AISTD differs from the ISTD in that it adjusts the color inconsistency between shadow and shadow-free images through image processing algorithms~\cite{Le2019Shadow}, where the color mismatch is caused by data acquisition.
For both ISTD and AISTD datasets, our model is trained on the training set and evaluated on the testing set. Similar to~\cite{Le_tpami21,liu2021shadow,liu2021from}, we use ground-truth shadow masks for training, and for testing, shadow masks are detected by BDRAR~\cite{zhu2018bidirectional} that is trained on SBU~\cite{vicente2016large} and AISTD~\cite{Le2019Shadow} datasets. 

\textbf{SRD}~\cite{qu2017deshadownet} contains 3,088 pairs of shadow and shadow-free images, and it is split into 2,680 training and 408 testing image pairs. Following~\cite{zhang2019shadowgan,Fu_2021_Auto}, we employ Otsu's algorithm with the difference between shadow and shadow-free images as initial masks to generate the shadow masks for training. In the inference phase, we utilize the public shadow masks predicted by DHAN~\cite{Cun_Pun_Shi_2020_DHAN} for testing and evaluation, similar to~\cite{hu2019direction,Cun_Pun_Shi_2020_DHAN,Fu_2021_Auto,jin2021dc,Chen_2021_CANet}.

\textbf{Video Shadow Removal}~\cite{le2019weakly} is an unseen dataset used to evaluate the generalization ability of models, which is composed of 8 challenging natural scenes. Each scene contains one video captured in a static environment with moving shadows. Following~\cite{le2019weakly,liu2021shadow,liu2021from}, we set the threshold to 40 and utilize it to split shadow and non-shadow pixels according to the intensity difference for evaluation. Additionally, we also employ the pre-trained BDRAR~\cite{zhu2018bidirectional} to generate the shadow mask of each frame for testing.

\subsection{Implementation Details}
\label{subsec:implementation}
We implement the proposed CRFormer using PyTorch and all the experiments are conducted on a single NVIDIA GeForce GTX 2080Ti GPU card. During training, we set the image input size to 400 $\times$ 400 with random flipping and cropping. We use the Adam~\cite{Kingma2015AdamAM} optimizer with the first and second momentum set to 0.50 and 0.99 respectively where we set the initial learning rate to $2 \times 10^{-4}$ and halved every 50 epochs. We train our CRFormer for a total of 200 epochs with the mini-batch size set to 1.

During evaluation, we record the root mean squared error (RMSE\footnote{As in~\cite{liu2021from,Fu_2021_Auto,jin2021dc,Le_tpami21}, RMSE is actually computed by the mean absolute error (MAE).}) metric for all datasets which performs in LAB color space. Furthermore, we employ the learned perceptual image patch similarity (LPIPS)~\cite{Zhang2018LPIPS} on the AISTD~\cite{Le2019Shadow} dataset to reflect the perceptual quality of the shadow-removal results.

\subsection{Comparisons with SOTA Methods}
\label{subsec:sota}

\begin{table}[t]\small
    \centering
    \caption{
    Shadow removal results of our CRFormer compared to state-of-the-art shadow-removal methods on the ISTD~\cite{wang2018stacked} testing set. S, NS and All represent the shadow region, non-shadow region and whole image, respectively. RMSE is the lower the better. `*' indicates that the result is directly cited from the original publication.
    }
    \renewcommand\arraystretch{1}
    \setlength{\tabcolsep}{3.0mm}{
    	{\begin{tabular}{l|c|c|c}
    		\toprule[1.0pt]
    		\multirow{2}{*}{Method}&\multicolumn{1}{c|}{\textbf{S} }&\multicolumn{1}{c|}{\textbf{NS}}& \multicolumn{1}{c}{\textbf{All}}\\ 
    		&RMSE&RMSE&RMSE\\ 
    		\midrule[0.5pt]
    		Input Image & 40.28 & 4.76 & 14.11\\
    		\midrule[0.5pt]
    		Gong \& Cosker~\cite{gong2014interactive} & 14.98 & 7.29 & 8.53\\
    		
    		Mask-ShadowGAN~\cite{hu2019mask} & 12.67 & 6.68 & 7.41\\
    
    		ST-CGAN~\cite{wang2018stacked} & 10.33 & 6.93 & 7.47\\
    		
    		DSC~\cite{hu2019direction} & 9.76 & 6.14 & 6.67\\
    
    		DHAN~\cite{Cun_Pun_Shi_2020_DHAN} & 8.14 & 6.04 & 6.37\\
    
    		AEFNet~\cite{Fu_2021_Auto} & 7.77 & \textbf{5.56} & \textbf{5.92}\\
    		
    		CANet*~\cite{Chen_2021_CANet} & 8.86 & 6.07 & 6.15\\
    		
            \midrule[0.5pt]
    		CRFormer~(Ours) & \textbf{7.32} & {5.82} & {6.07}\\
    		
    		\bottomrule[1.0pt]
    \end{tabular}}}
    \vspace{-.3cm}
    \label{tab:istd}
\end{table}

\begin{figure*}[htbp]\small
    \centering
	\begin{tabular}{ccccccc}
		\hspace{-.2cm}
		\includegraphics[width=.138\textwidth, height=0.1\textwidth]{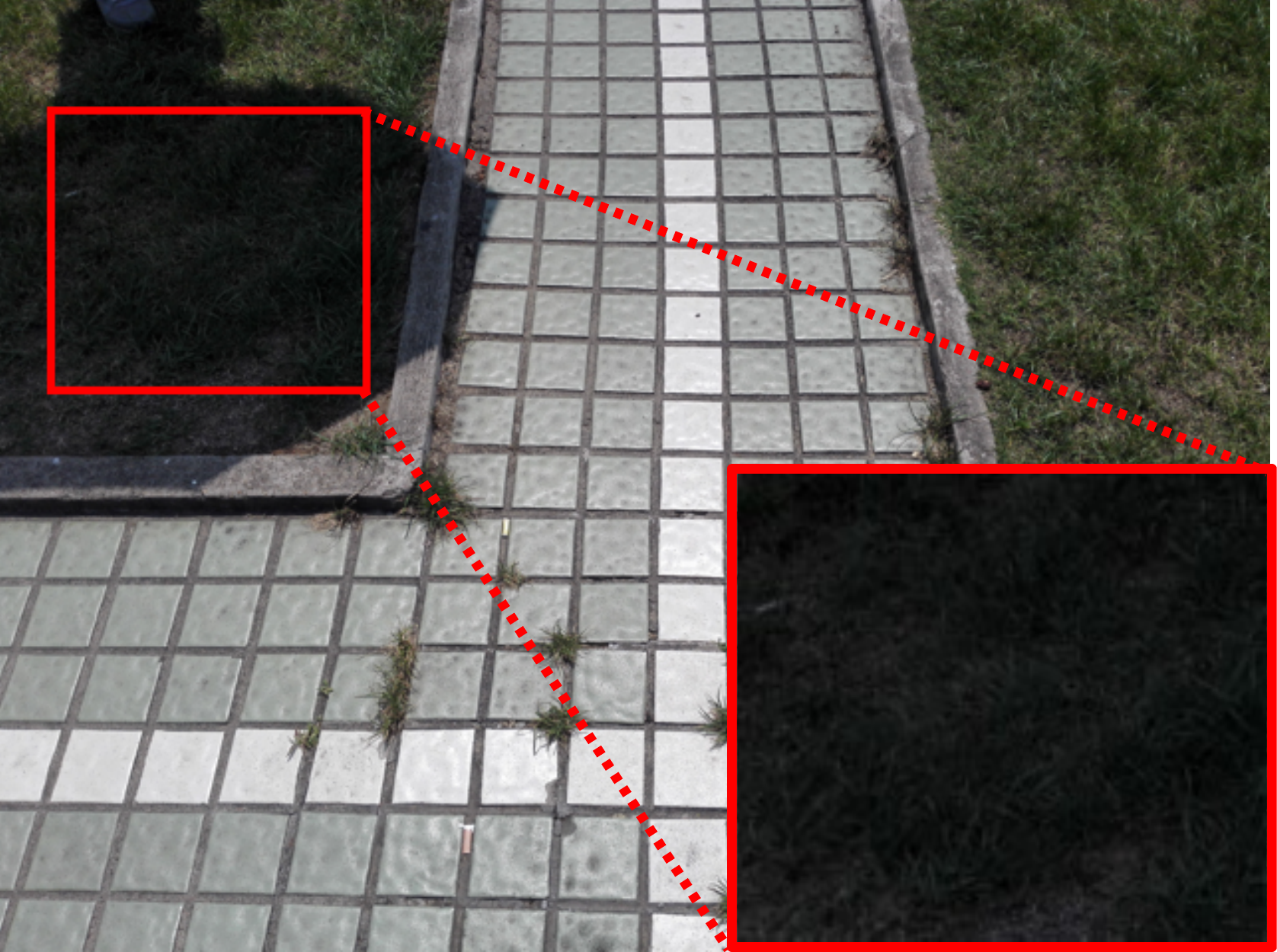} & \hspace{-.45cm}
		\includegraphics[width=.138\textwidth, height=0.1\textwidth]{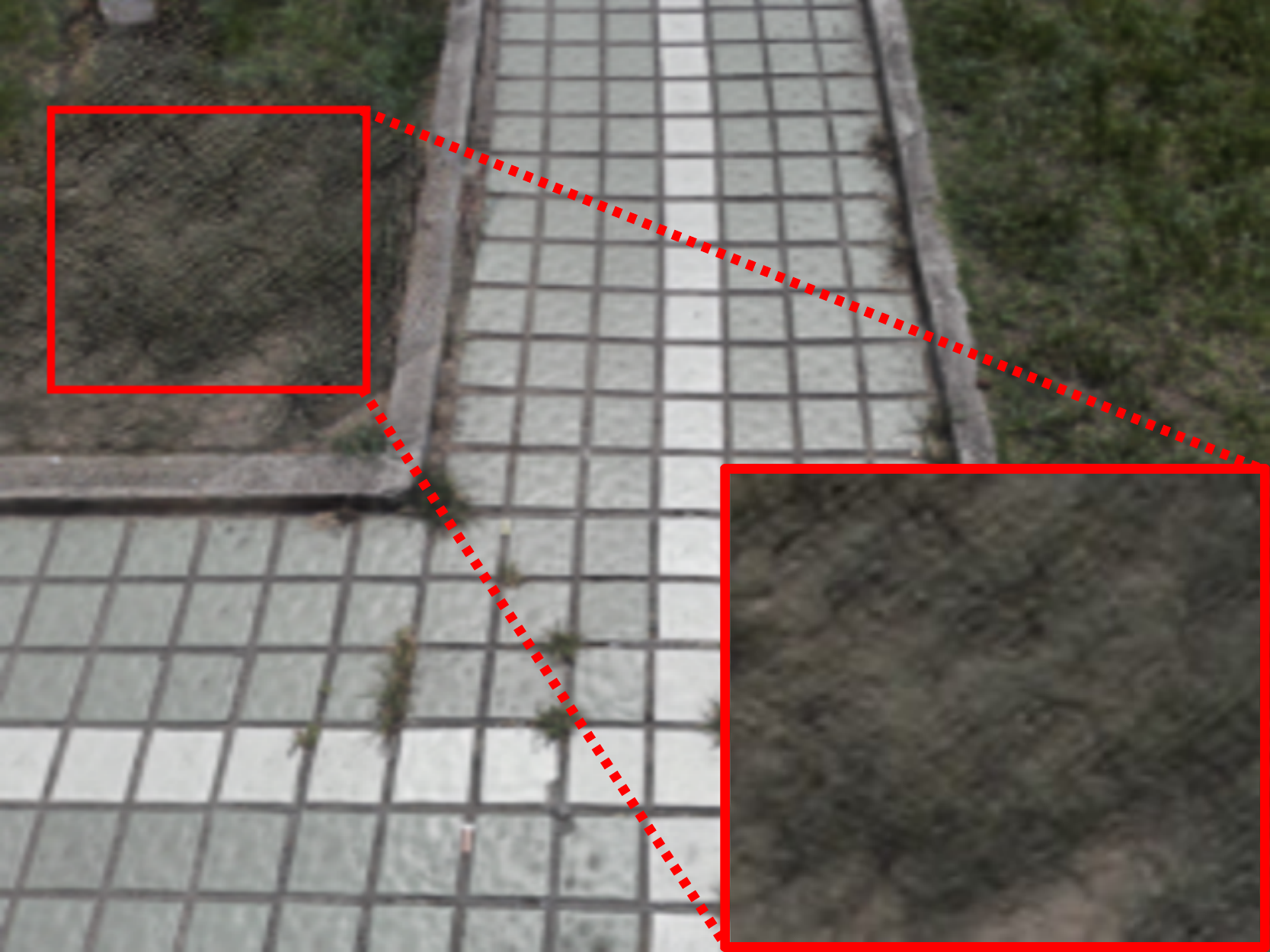} & \hspace{-.45cm}
		\includegraphics[width=.138\textwidth, height=0.1\textwidth]{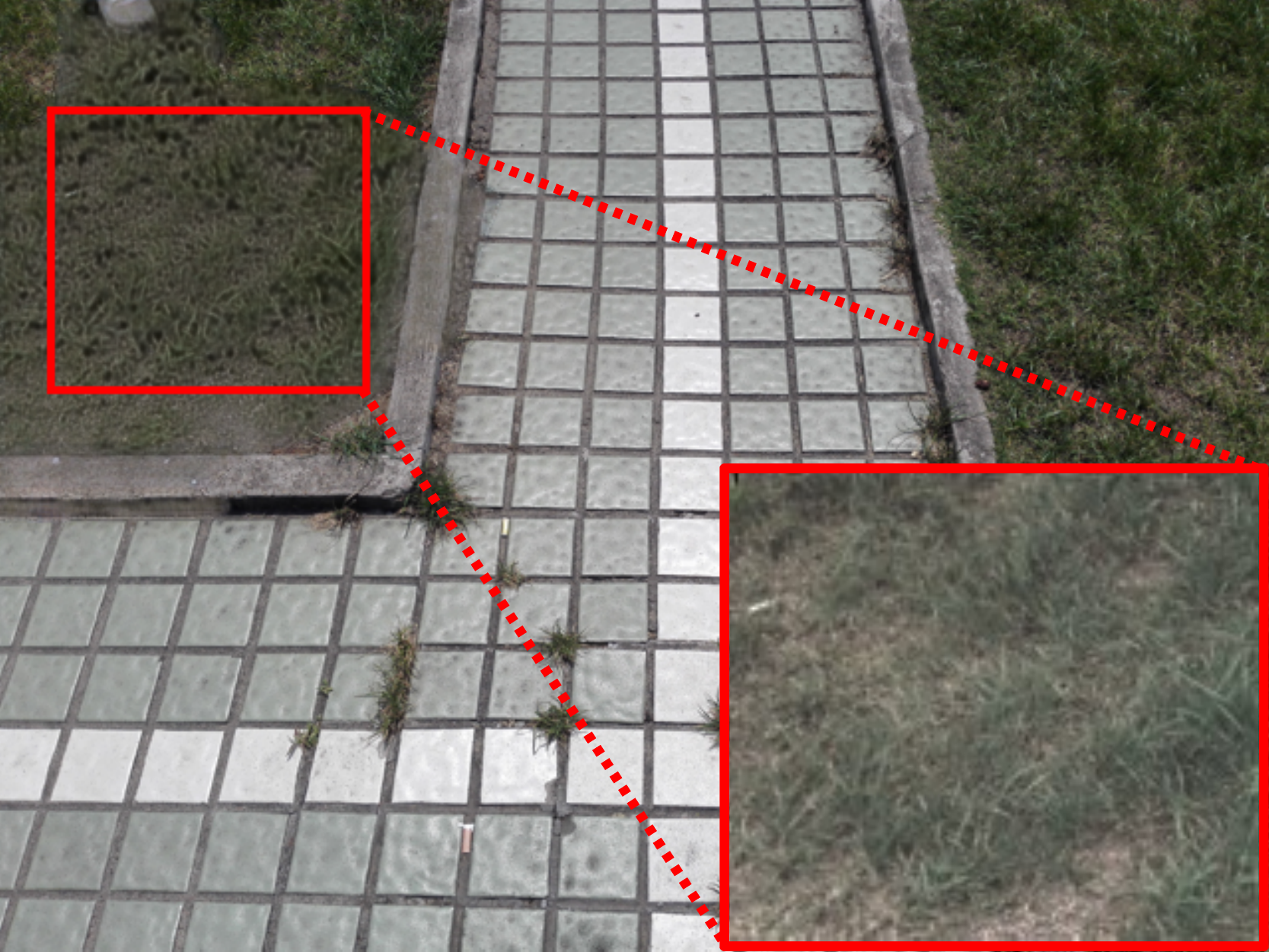}   & \hspace{-.45cm}
		\includegraphics[width=.138\textwidth, height=0.1\textwidth]{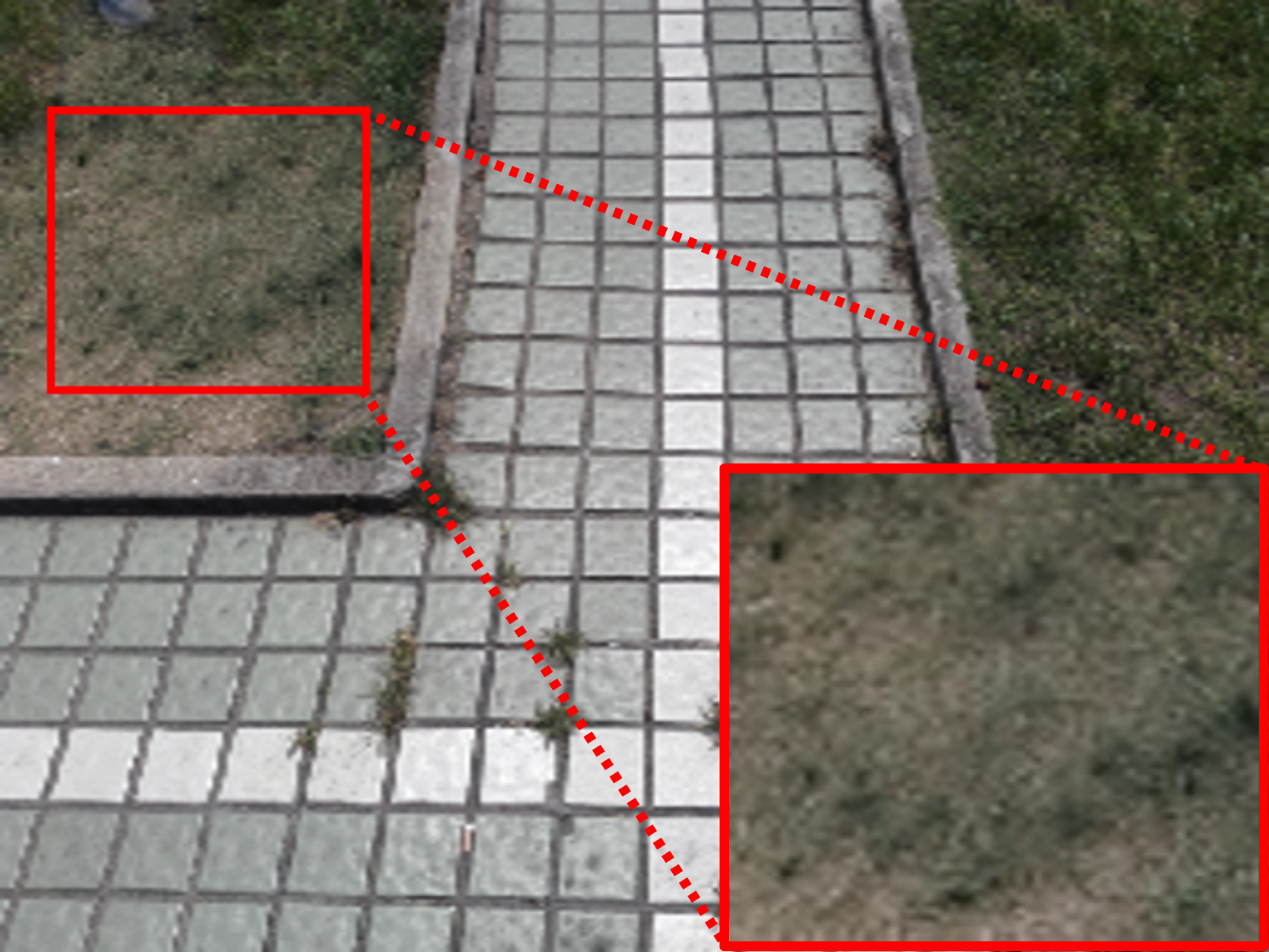} & \hspace{-.45cm}
		\includegraphics[width=.138\textwidth, height=0.1\textwidth]{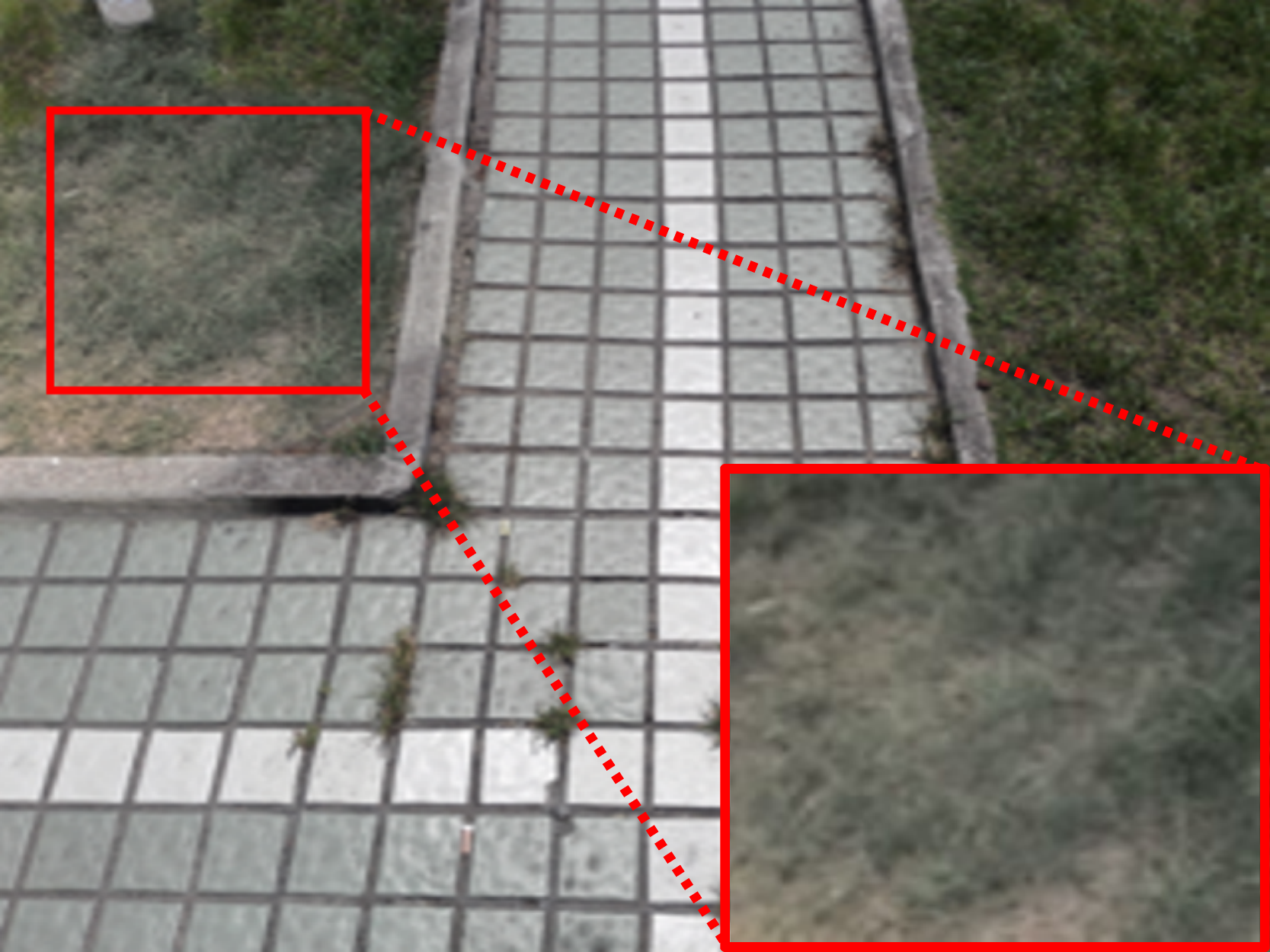} & \hspace{-.45cm}
		\includegraphics[width=.138\textwidth, height=0.1\textwidth]{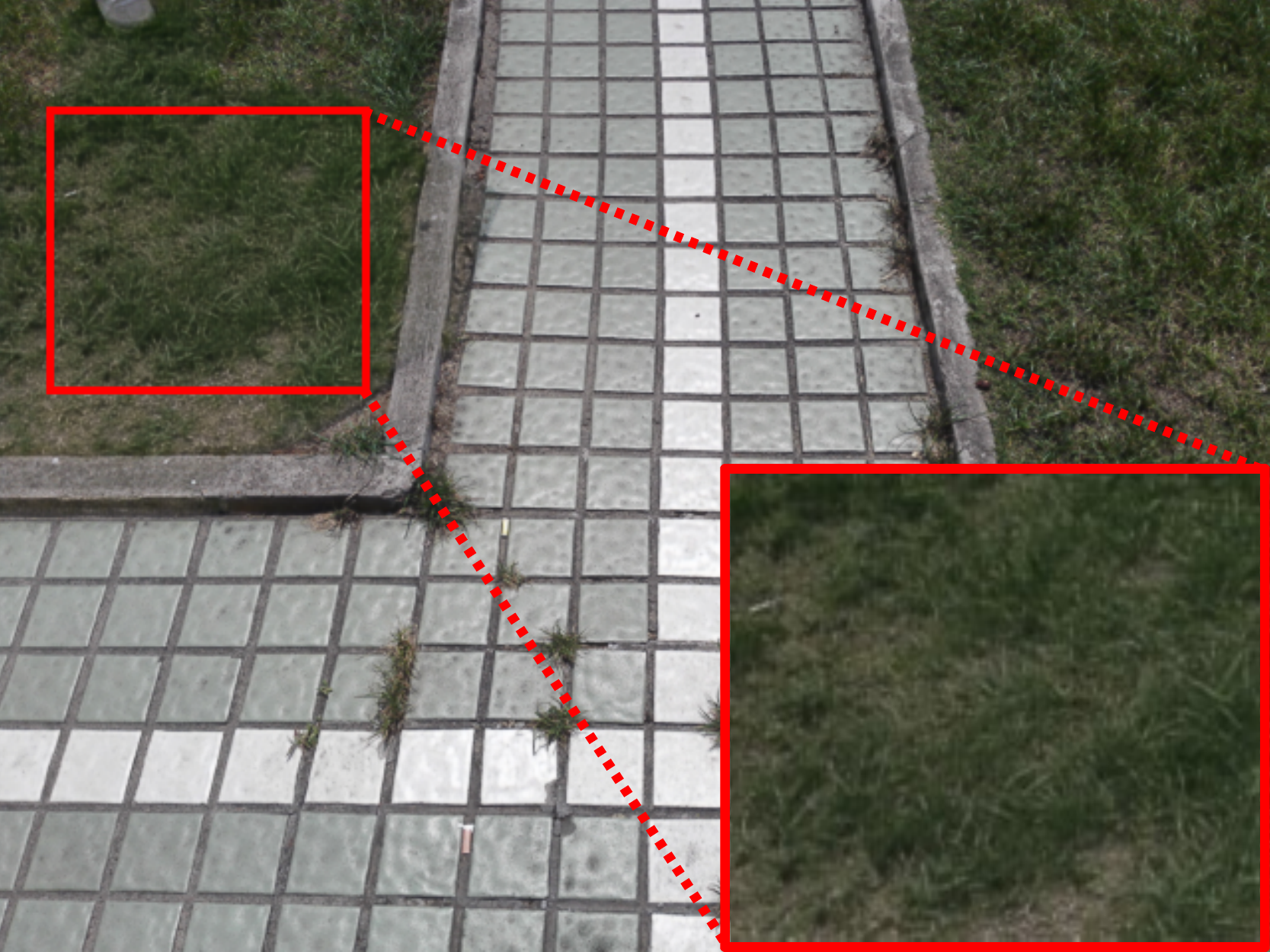} & \hspace{-.45cm}
		\includegraphics[width=.138\textwidth, height=0.1\textwidth]{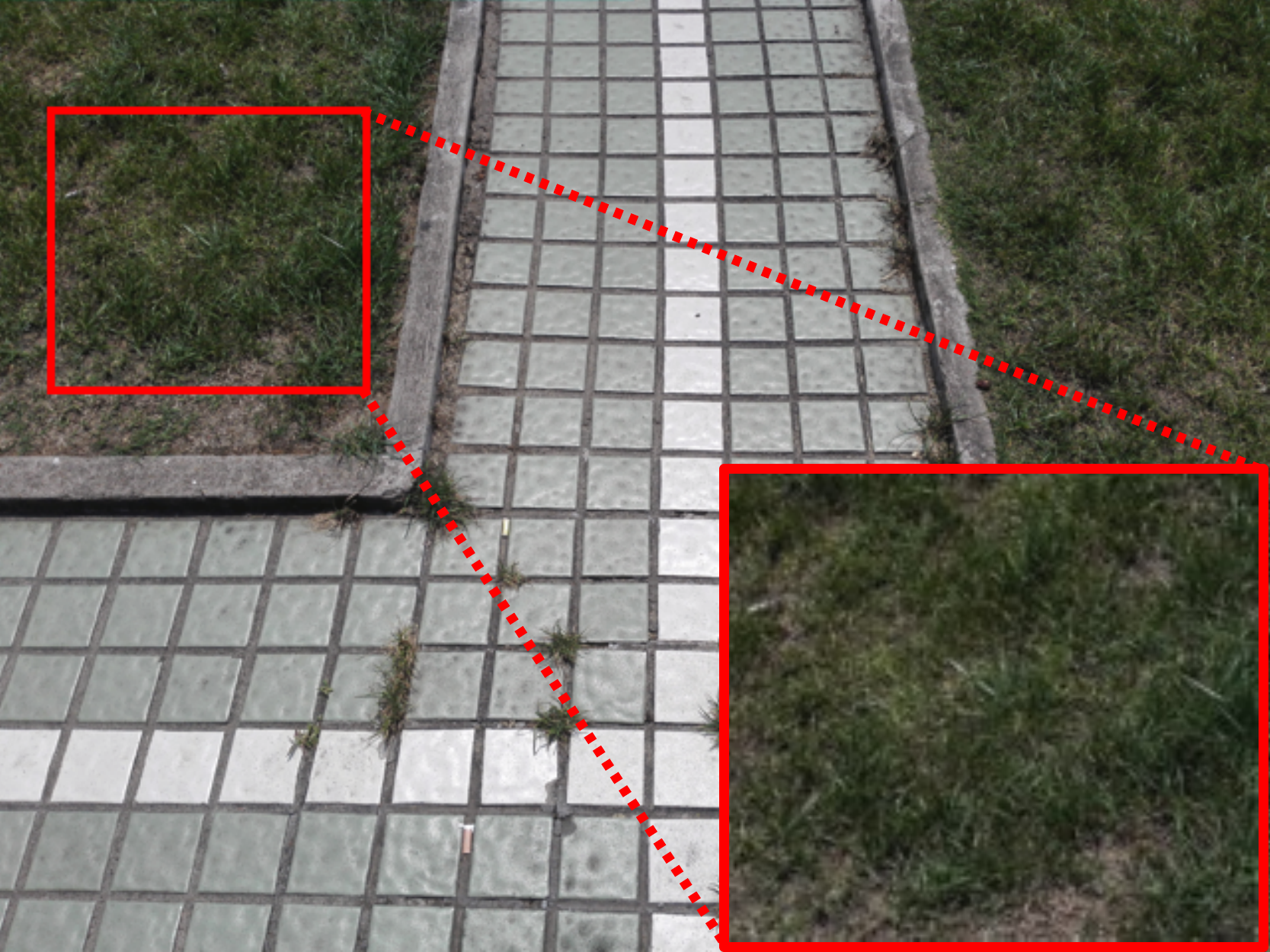}\vspace{-.04cm} \\
		\hspace{-.2cm}
		\includegraphics[width=.138\textwidth, height=0.1\textwidth]{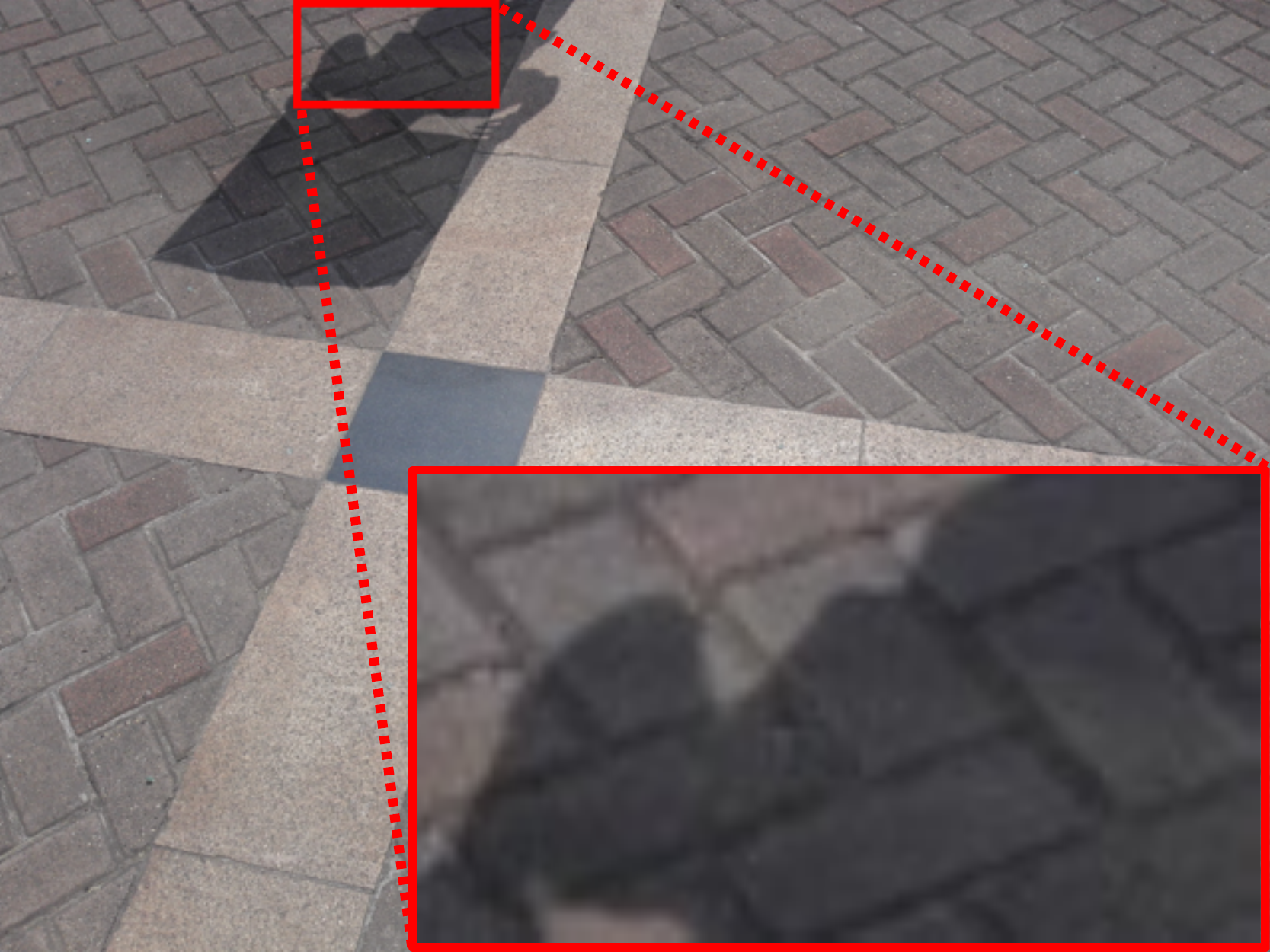} & \hspace{-.45cm}
		\includegraphics[width=.138\textwidth, height=0.1\textwidth]{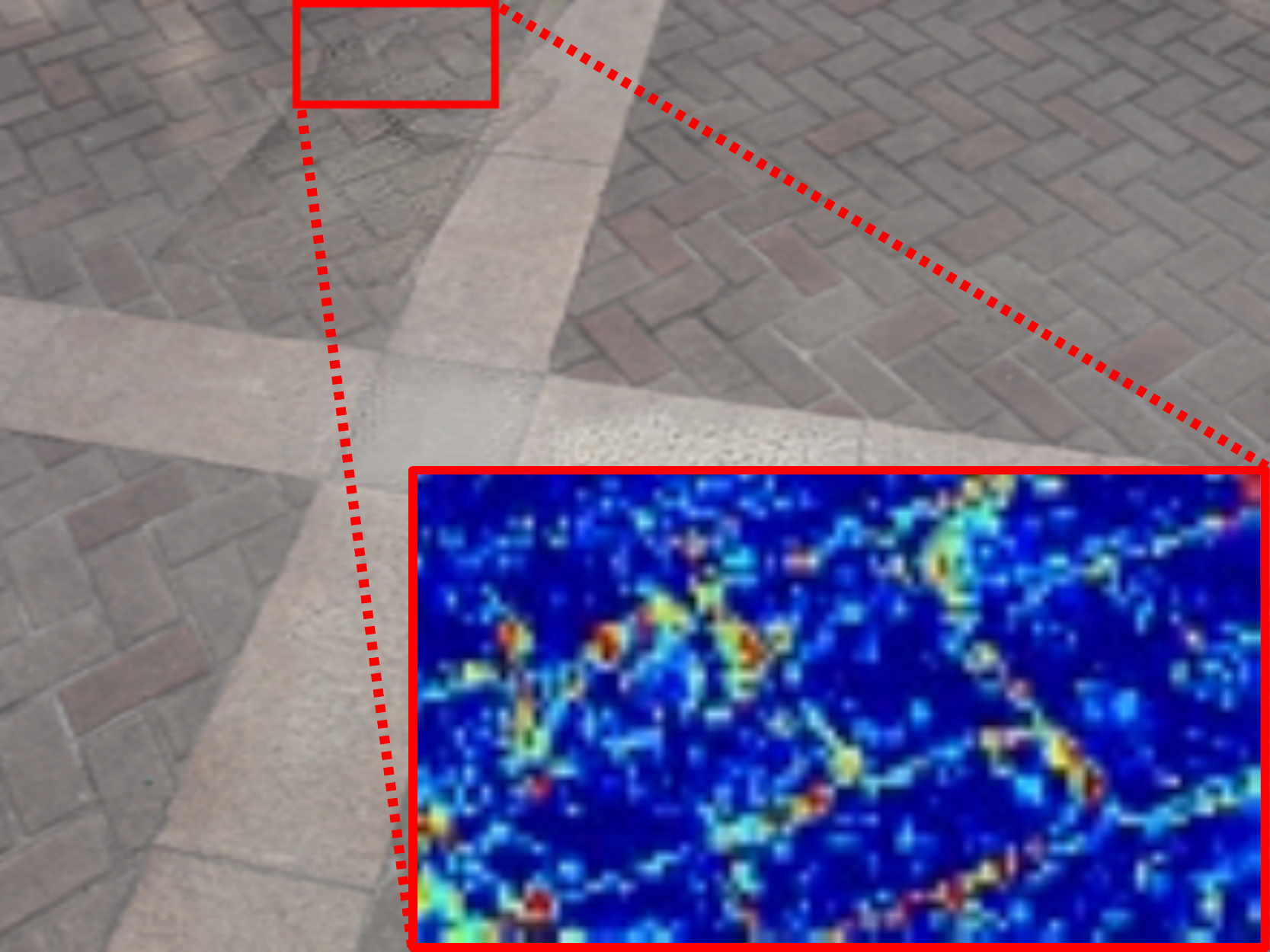} & \hspace{-.45cm}
		\includegraphics[width=.138\textwidth, height=0.1\textwidth]{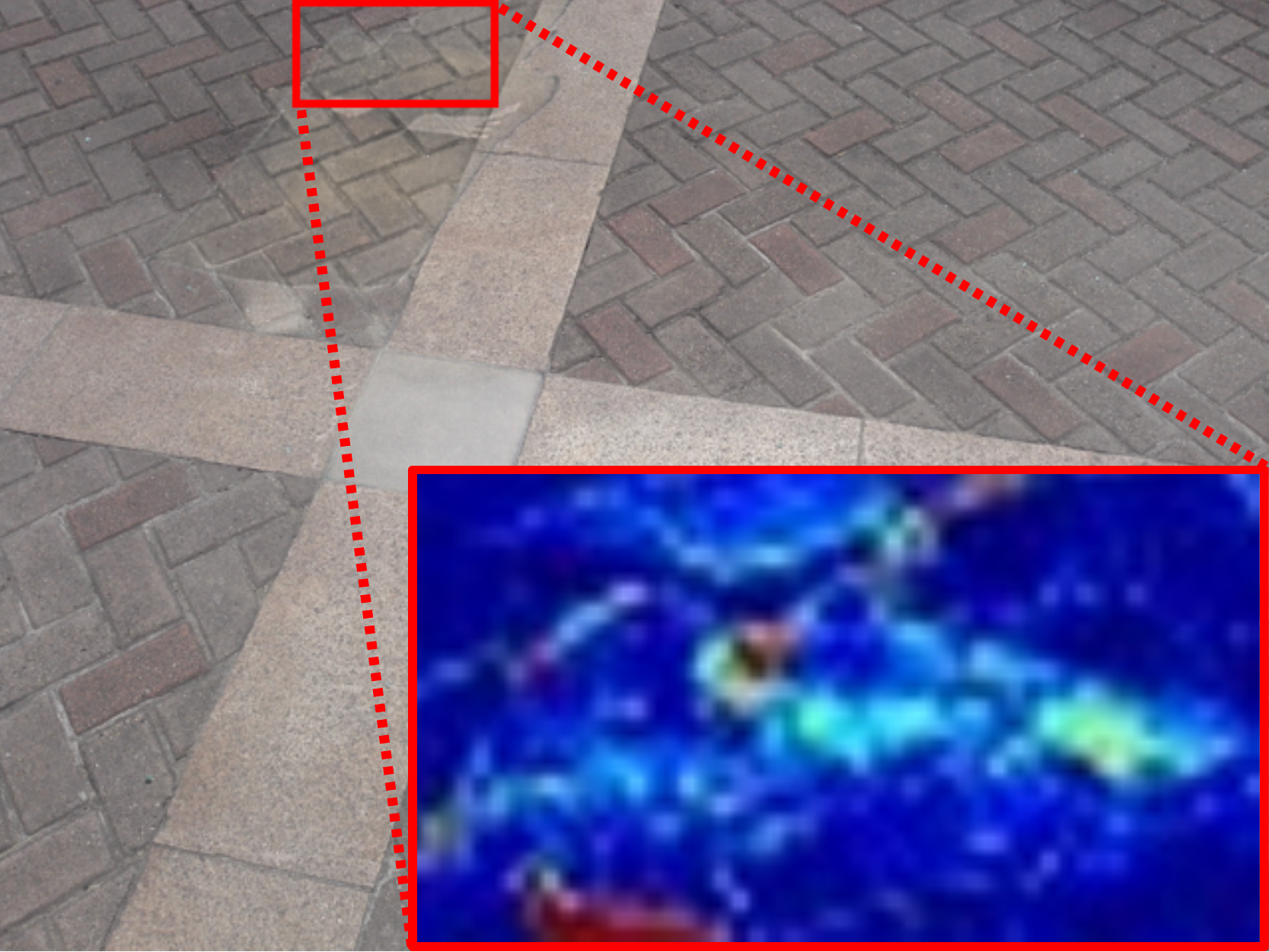}   & \hspace{-.45cm}
		\includegraphics[width=.138\textwidth, height=0.1\textwidth]{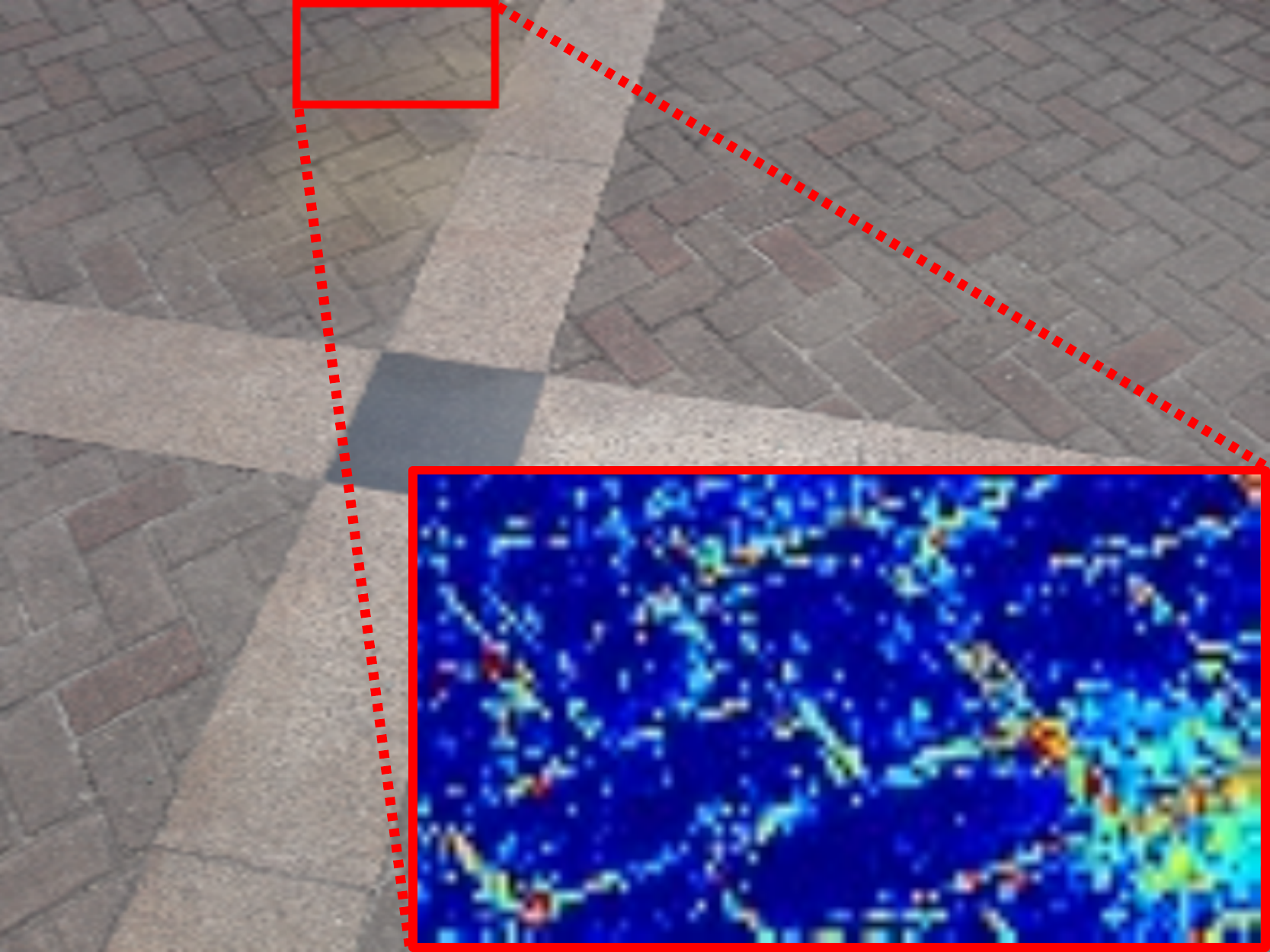} & \hspace{-.45cm}
		\includegraphics[width=.138\textwidth, height=0.1\textwidth]{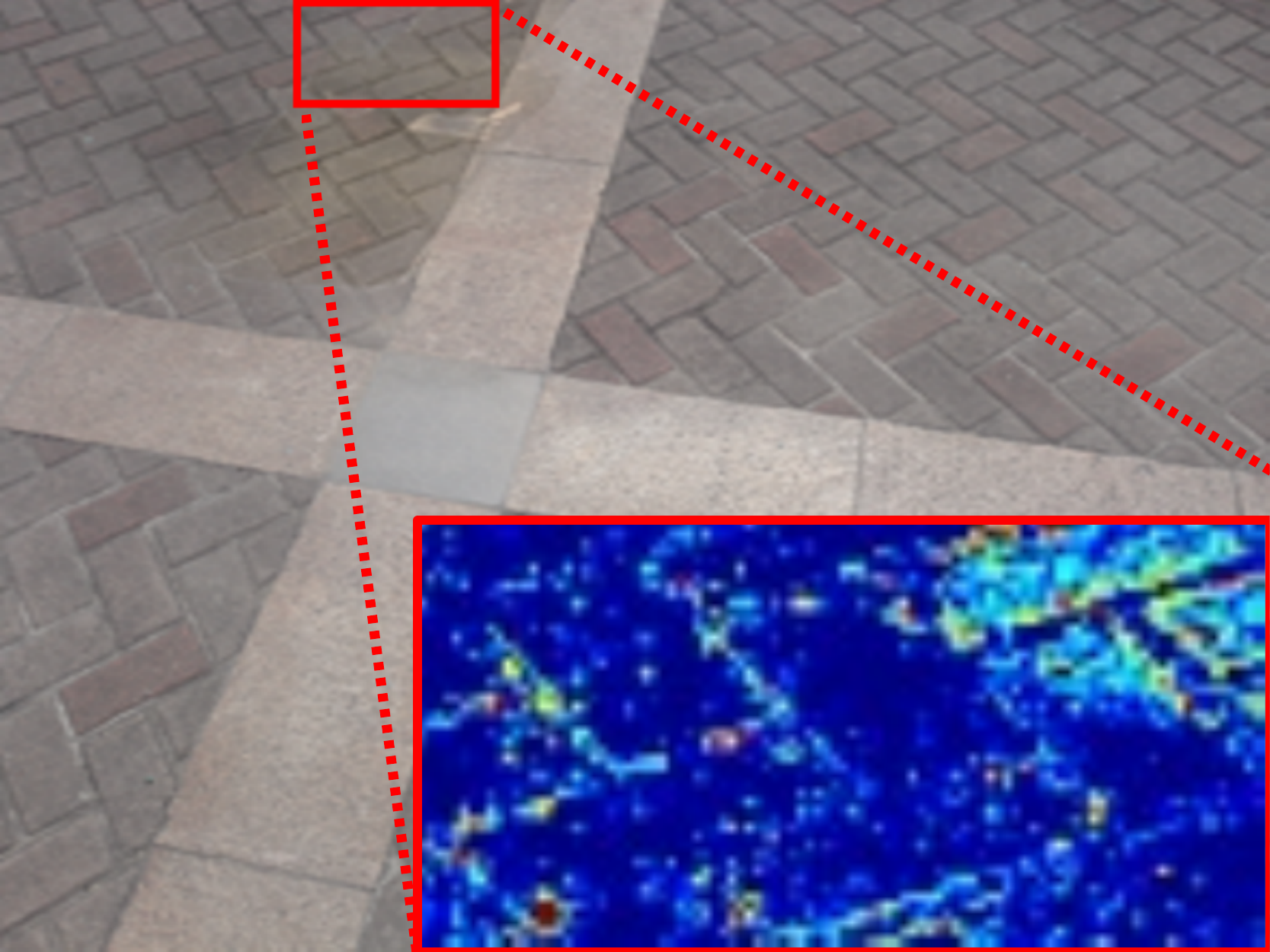} & \hspace{-.45cm}
		\includegraphics[width=.138\textwidth, height=0.1\textwidth]{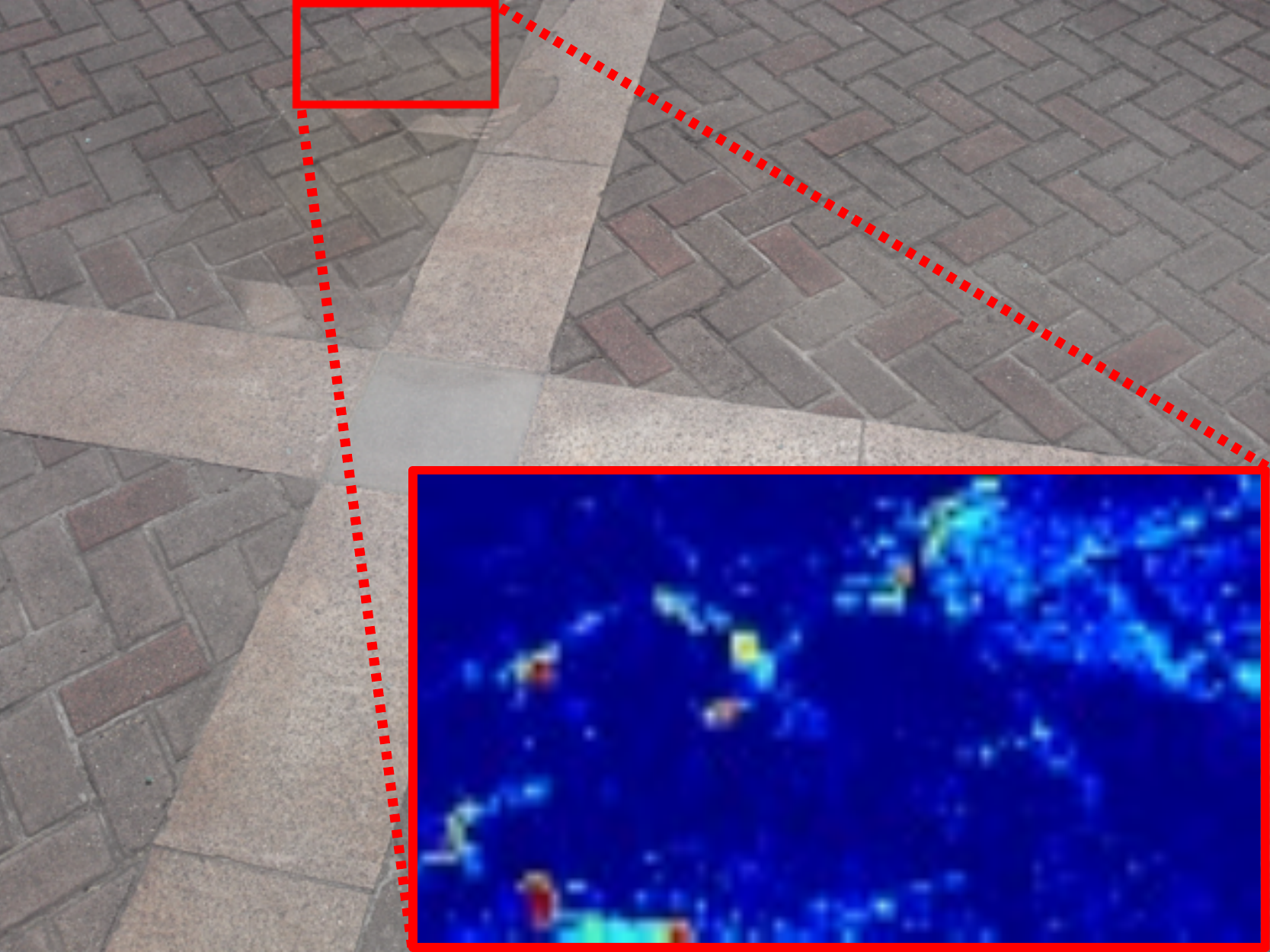} & \hspace{-.45cm}
		\includegraphics[width=.138\textwidth, height=0.1\textwidth]{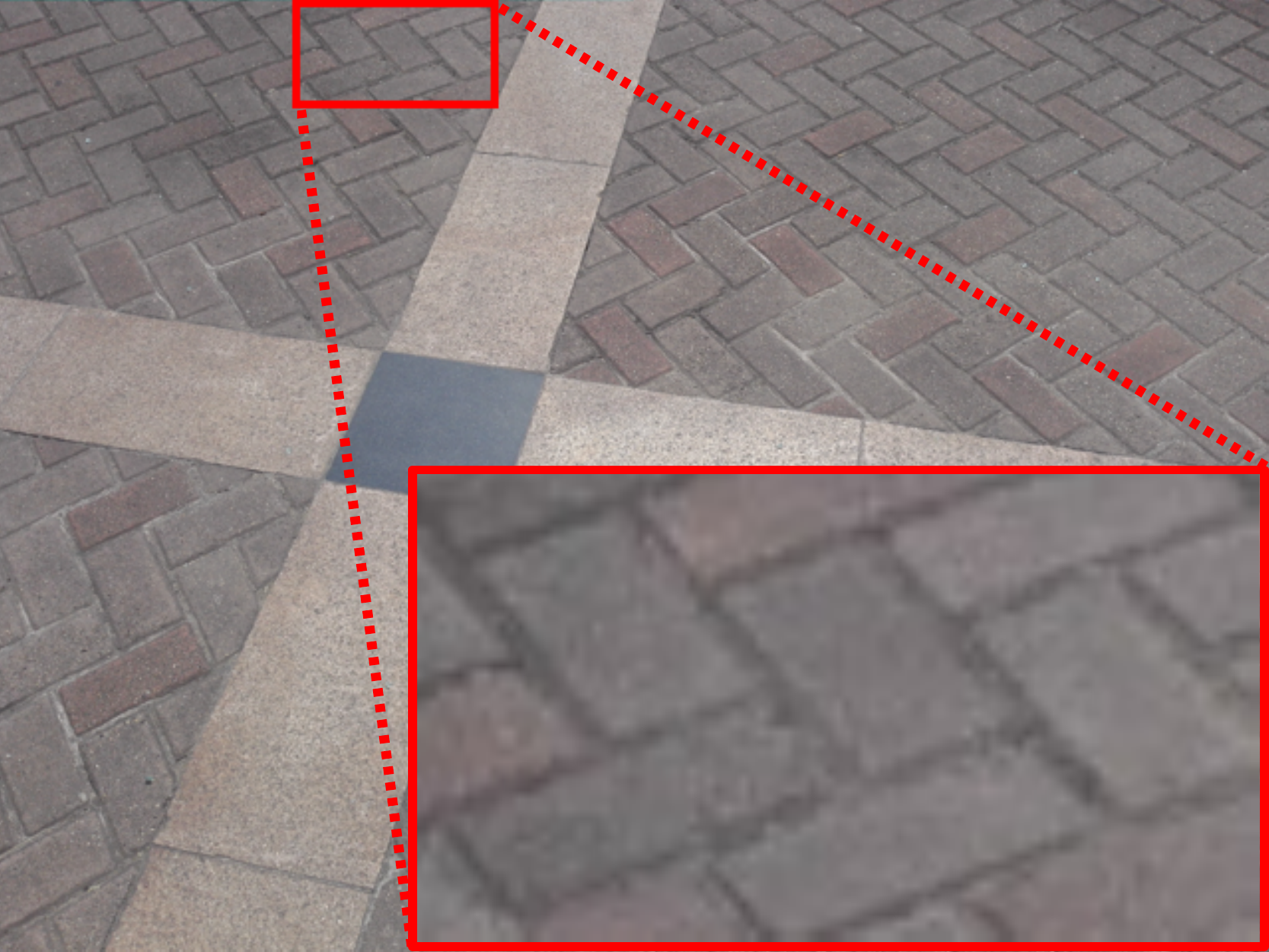}\vspace{-.04cm} \\
		\hspace{-.2cm}  Input & \hspace{-.45cm} {DC~\cite{Le2019Shadow}} & \hspace{-.45cm}  G2R~\cite{liu2021from} &\hspace{-.45cm} {AEFNet~\cite{Fu_2021_Auto}} &\hspace{-.45cm} {SP+M+I-Net~\cite{Le_tpami21}} &\hspace{-.45cm}  Ours &\hspace{-.45cm} GT \vspace{.04cm}\\
		\hspace{-.2cm}
		\includegraphics[width=.138\textwidth, height=0.1\textwidth]{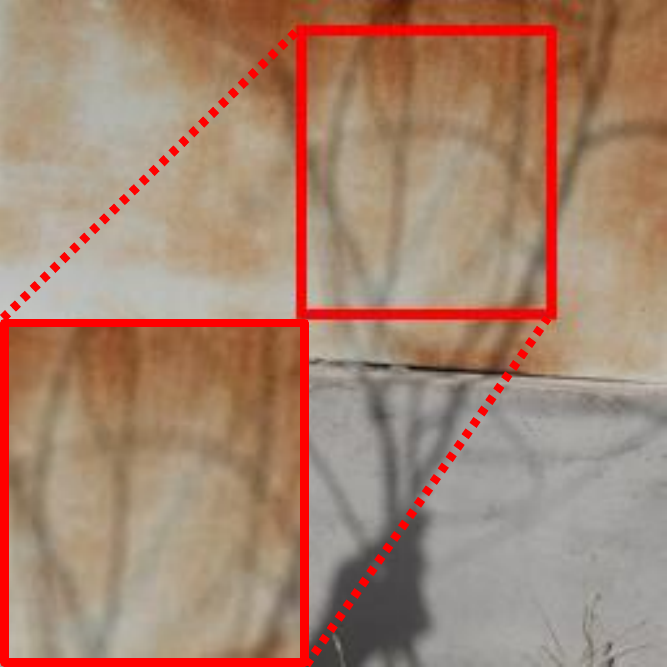} & \hspace{-.45cm} 
		\includegraphics[width=.138\textwidth, height=0.1\textwidth]{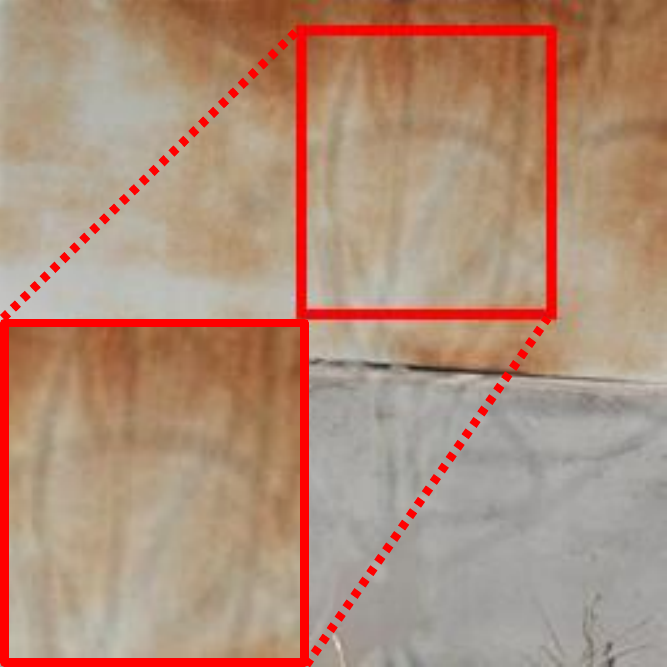} & \hspace{-.45cm}
		\includegraphics[width=.138\textwidth, height=0.1\textwidth]{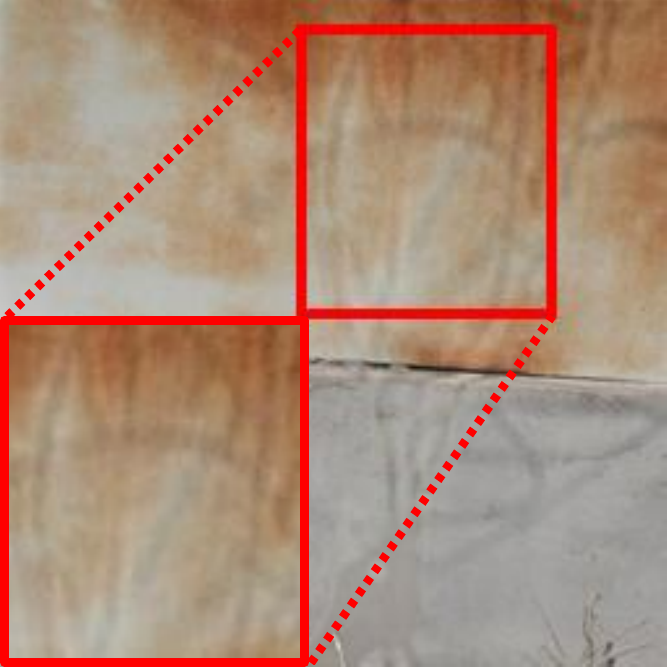} & \hspace{-.45cm}
		\includegraphics[width=.138\textwidth, height=0.1\textwidth]{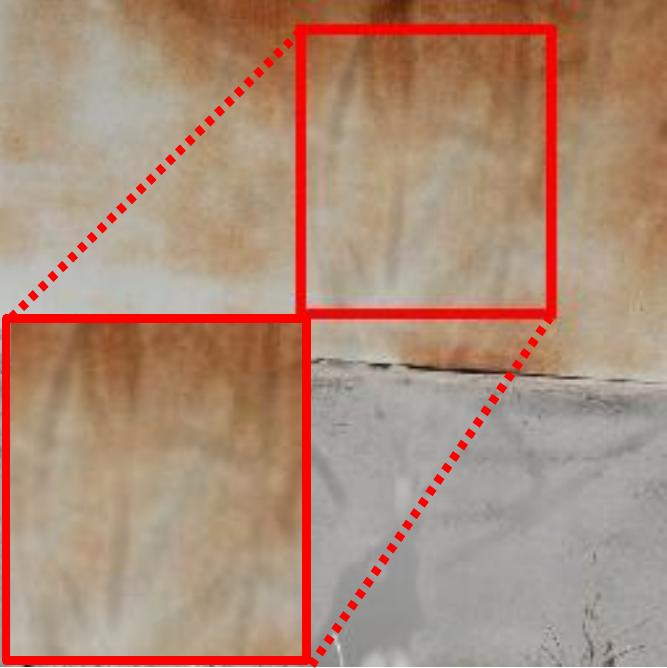} & \hspace{-.45cm}
		\includegraphics[width=.138\textwidth, height=0.1\textwidth]{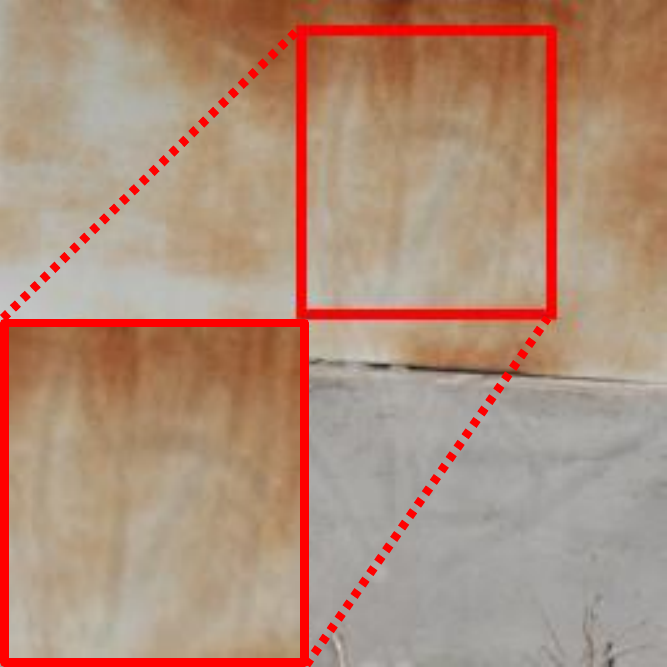} & \hspace{-.45cm}
		\includegraphics[width=.138\textwidth, height=0.1\textwidth]{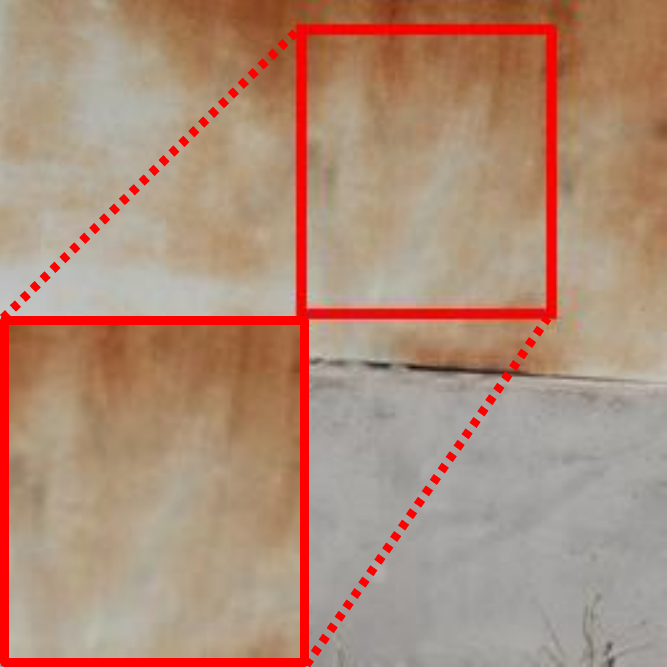} & \hspace{-.45cm}
		\includegraphics[width=.138\textwidth, height=0.1\textwidth]{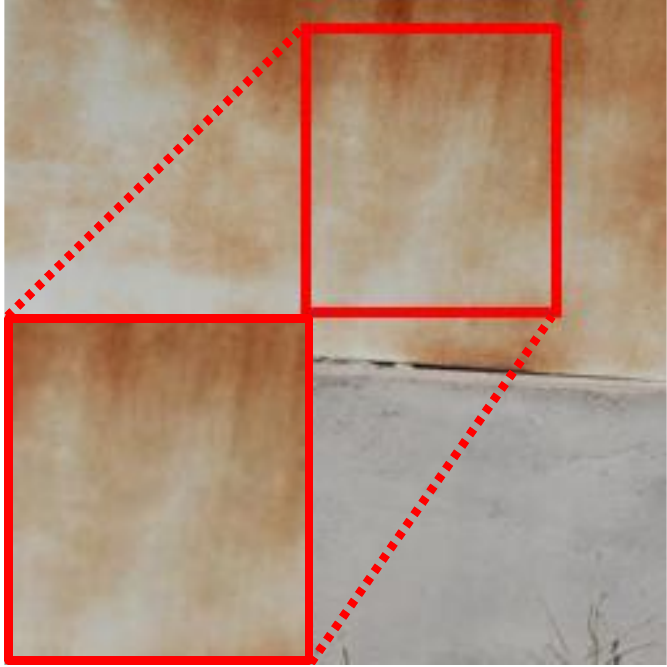}\vspace{-.04cm} \\
		\hspace{-.2cm}
		\includegraphics[width=.138\textwidth, height=0.1\textwidth]{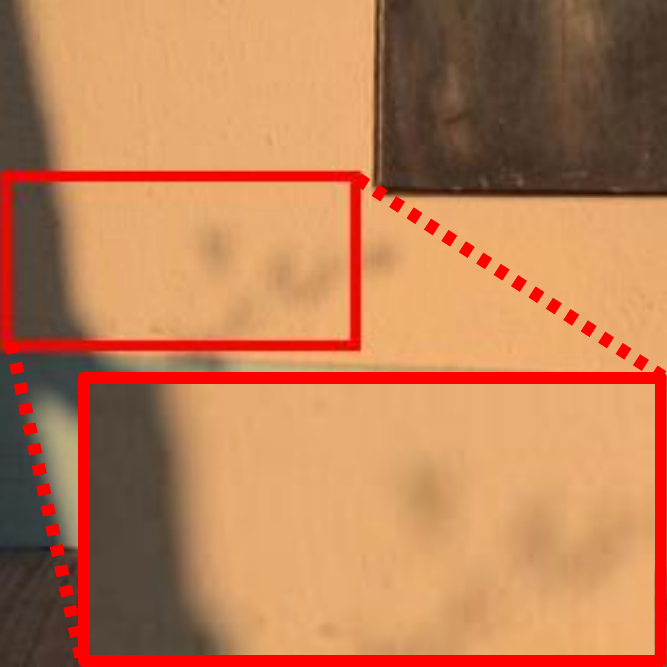} & \hspace{-.45cm} 
		\includegraphics[width=.138\textwidth, height=0.1\textwidth]{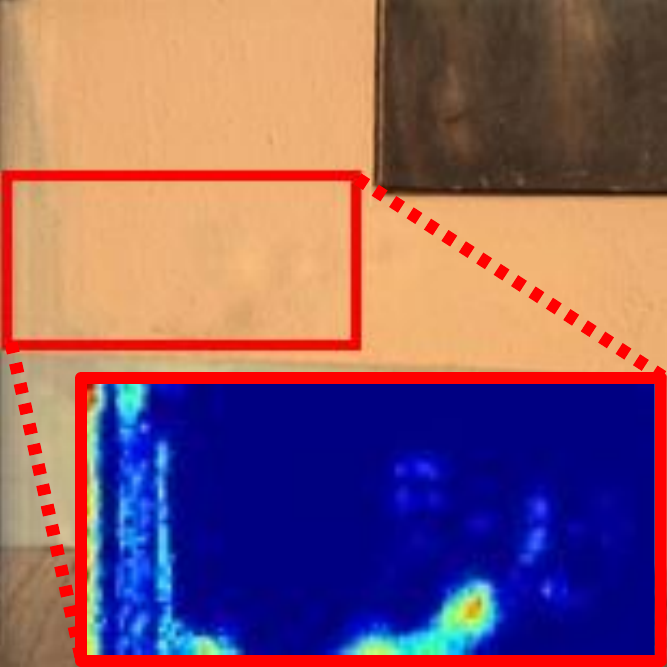} & \hspace{-.45cm}
		\includegraphics[width=.138\textwidth, height=0.1\textwidth]{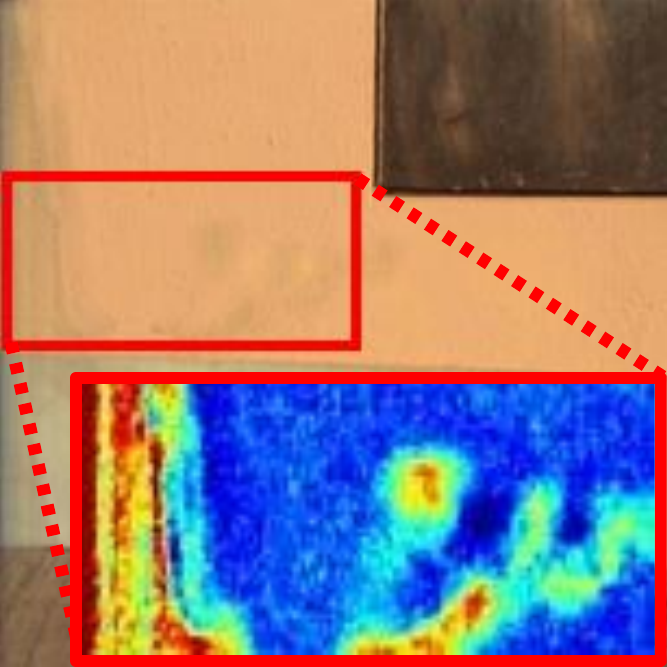} & \hspace{-.45cm}
		\includegraphics[width=.138\textwidth, height=0.1\textwidth]{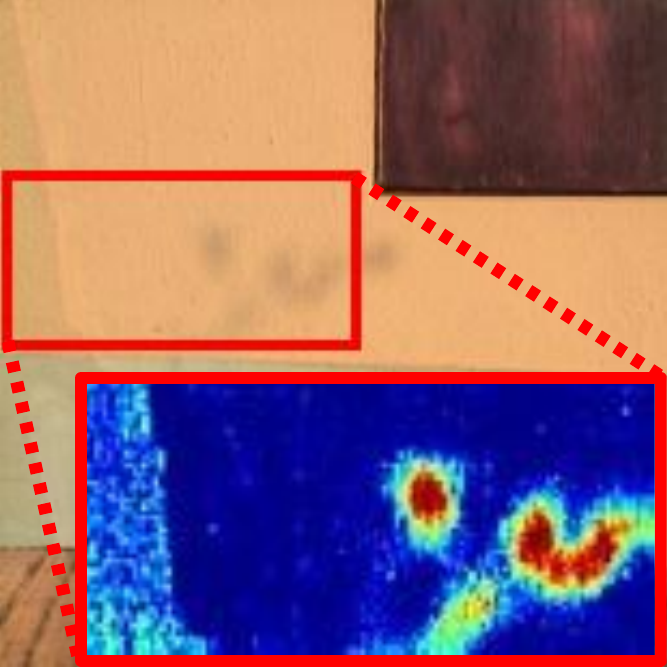} & \hspace{-.45cm}
		\includegraphics[width=.138\textwidth, height=0.1\textwidth]{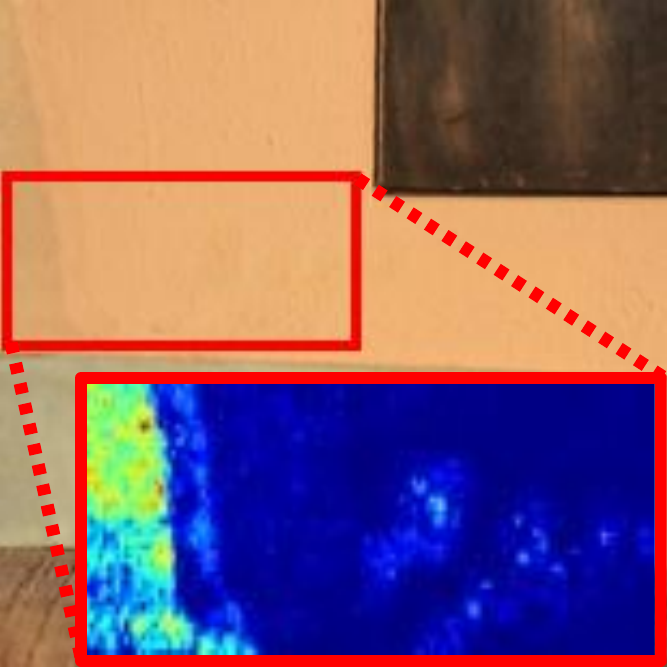} & \hspace{-.45cm}
		\includegraphics[width=.138\textwidth, height=0.1\textwidth]{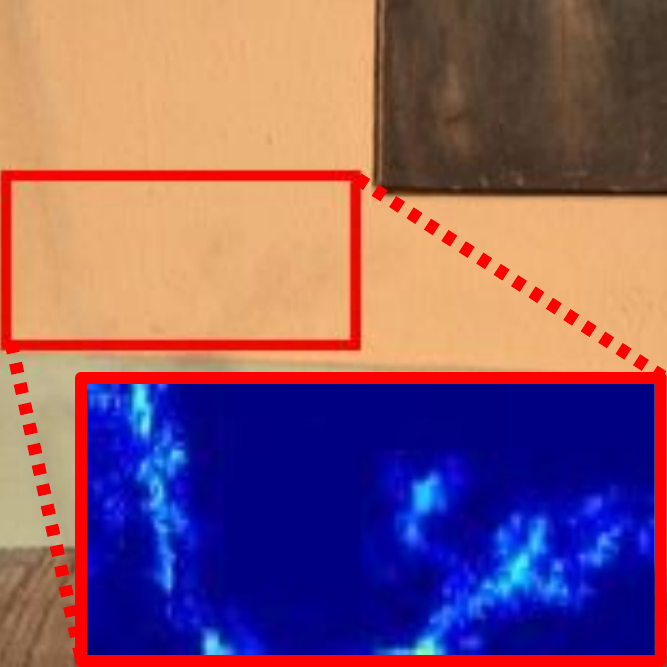} & \hspace{-.45cm}
		\includegraphics[width=.138\textwidth, height=0.1\textwidth]{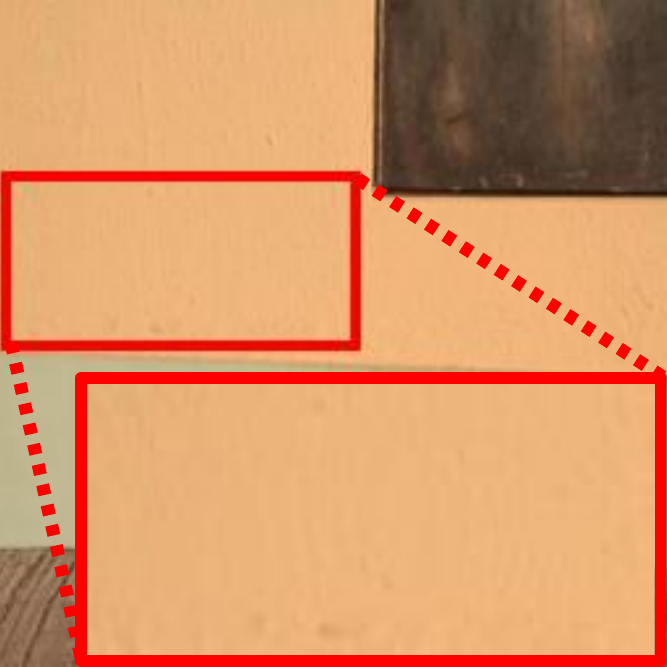}\vspace{-.0cm} \\
		\hspace{-.2cm}  Input & \hspace{-.45cm} 
		{DeshadowNet~\cite{qu2017deshadownet}} &\hspace{-.45cm}
		{DSC~\cite{hu2019direction}}&\hspace{-.45cm}
		{AEFNet~\cite{Fu_2021_Auto}} &\hspace{-.45cm} {DC~\cite{jin2021dc}} &\hspace{-.45cm}  Ours &\hspace{-.45cm} GT \\
	\end{tabular}
	\vspace{-4pt}
    \caption{
    Qualitative comparisons with different methods on the AISTD~\cite{Le2019Shadow} (top three rows) and SRD~\cite{qu2017deshadownet} (bottom two rows) datasets. Heatmaps are used to show the difference between the shadow-removal results and the ground truth.
    }
    \label{fig:3}
	\vspace{-.2cm}
\end{figure*}

\textbf{On ISTD and AISTD.} For ISTD~\cite{wang2018stacked}, we compare against state-of-the-art methods including Gong \& Cosker~\cite{gong2014interactive}, Mask-ShadowGAN \cite{hu2019mask}, ST-CGAN~\cite{wang2018stacked}, DSC~\cite{hu2019direction}, DHAN~\cite{Cun_Pun_Shi_2020_DHAN}, AEFNet~\cite{Fu_2021_Auto} and CANet~\cite{Chen_2021_CANet}. Table~\ref{tab:istd} shows the quantitative results of our method and other competitors. The results show that our method achieves the best shadow removal performance in shadow regions and produces competitive results in non-shadow regions and the whole image. Compared with the second best~\cite{Fu_2021_Auto}, the proposed method achieves a gain of 5.8\% on the RMSE score in the shadow region. It also outperforms the recent patch-based supervised method CANet with a 17.4\% RMSE reduction in the shadow region which proves that considering all pixel-level features from non-shadow regions in our work can capture adequate contextual information of non-shadow regions to restore the pixel intensity of shadow regions, yielding better shadow-removal performance. 

\begin{table}[htbp]\small
    \centering
    \caption{
    Shadow removal results of our CRFormer compared to state-of-the-art shadow-removal methods on the AISTD~\cite{Le2019Shadow} testing set. LPIPS is the lower the better.
    }
    \renewcommand\arraystretch{1.1}
    \setlength{\tabcolsep}{0.40mm}{
    	{\begin{tabular}{l|c|c|c|cc}
    		\toprule[1.0pt]
    		\multirow{2}{*}{Scheme}&\multirow{2}{*}{Method}&\multicolumn{1}{c|}{\textbf{S} }&\multicolumn{1}{c|}{\textbf{NS}}& \multicolumn{2}{c}{\textbf{All}}\\ 
    		&&RMSE&RMSE&RMSE&LPIPS\\ 
    		\midrule[0.5pt]
    		\multirow{3}{*}{\shortstack{Un- \\ supervised}} 
    		& Mask-ShadowGAN~\cite{wang2018stacked} &9.9 & 3.8 & 4.8 & 0.095 \\
    
    		& LG-ShadowNet~\cite{liu2021shadow} &9.7& 3.4 & 4.4&0.103\\
    		
    		& DC-ShadowNet~\cite{jin2021dc} &10.4& 3.6 & 4.7&0.170\\
            \midrule[0.5pt]
    		\multirow{3}{*}{\shortstack{Weakly- \\ supervised}} & Gong \& Cosker~\cite{gong2014interactive} &13.3 & \textbf{2.6} & 4.3 &0.086\\
    		& Param+M+D-Net~\cite{le2020from} & 9.7 & 2.9 & 4.1 &0.086\\
    
    		& G2R-ShadowNet~\cite{liu2021from} & 8.8 & 2.9 & 3.9 &0.096\\
    		\midrule[0.5pt]
    
    		\multirow{5}{*}{\shortstack{Fully- \\ supervised}} & ST-CGAN~\cite{wang2018stacked} &13.4 & 7.9 & 8.6 &0.150\\
    
    		& SP+M-Net~\cite{Le2019Shadow} & 7.9 & {2.8} & 3.6 & 0.085\\
    
    		& AEFNet~\cite{Fu_2021_Auto} & 6.5 & 3.8 & 4.2 & 0.106\\
    		
    		& SP+M+I-Net~\cite{Le_tpami21} & 6.0 & 3.1 & 3.6 & 0.092\\
    
    		& CRFormer~(Ours) & \textbf{5.9} & 2.9 & \textbf{3.4} & \textbf{0.068} \\
    		
    		\bottomrule[1.0pt]
    \end{tabular}}}
    \vspace{-0.3cm}
    \label{tab:aistd}
\end{table}

For AISTD~\cite{Le2019Shadow}, we compare our model with \textit{unsupervised methods} including Mask-ShadowGAN~\cite{wang2018stacked}, LG-ShadowNet~\cite{liu2021shadow}, and DC-ShadowNet~\cite{jin2021dc}, \textit{weakly-supervised methods} including Gong \& Cosker~\cite{gong2014interactive}, Param+M+D-Net~\cite{le2020from}, and G2R-ShadowNet~\cite{liu2021from}, and \textit{supervised methods} including ST-CGAN~\cite{wang2018stacked}, SP+M-Net~\cite{Le2019Shadow}, AEFNet~\cite{Fu_2021_Auto}, and SP+M+I-Net~\cite{Le_tpami21}. As shown in Table~\ref{tab:aistd}, our method outperforms all state-of-the-art methods in both shadow regions and the entire image. Meanwhile, LPIPS scores also show consistent performance, which shows that the proposed method produces results with better visual perceptual quality. Specifically, our method achieves the best results compared to unsupervised and weakly-supervised methods, and it outperforms recent DC-ShadowNet and G2R-ShadowNet in shadow regions with RMSE being decreased by 43.3\% and 33.0\%, respectively. In addition, compared to supervised methods, our method still performs the best on the shadow region and the entire image.
It outperforms AEFNet with a 9.2\% RMSE reduction in the shadow region. Although SP+M+I-Net shows a close performance in shadow regions to ours, we get 5.6\% and 26.1\% gain over it in RMSE and LPIPS on the whole image, respectively.

Qualitative comparisons are shown in Fig.~\ref{fig:3}~(top three rows). We can see that GAN-based DC-ShadowNet and G2R-ShadowNet tend to produce blurry details in the shadow region due to data distribution requirements. In addition, AEFNet and SP+M+I-Net also reconstruct some unsatisfactory de-shadowed results (\textit{e.g.}, color distortion and artifacts) without fully leveraging the long-range information from the non-shadow regions. Conversely, our method produces visually pleasing shadow-free images by transferring adequate contextual information from non-shadow to shadow regions.

\begin{table}[htbp]\small
    \centering
    \caption{
    Shadow removal results of our CRFormer compared to state-of-the-art shadow-removal methods on the SRD~\cite{qu2017deshadownet} testing set. 
    }
    \renewcommand\arraystretch{1.}
    \setlength{\tabcolsep}{3.5mm}{
    {\begin{tabular}{l|c|c|c}
    	\toprule[1.0pt]
    	\multirow{2}{*}{Method}&\multicolumn{1}{c|}{\textbf{S}}&\multicolumn{1}{c|}{\textbf{NS}}& \multicolumn{1}{c}{\textbf{All}}\\
    	&RMSE&RMSE&RMSE\\ 
    	\midrule[0.5pt]
    DeshadowNet~\cite{qu2017deshadownet}&11.78 & 4.84&6.64\\
    DSC~\cite{hu2019direction} & 10.89 & 4.99 & 6.23 \\
    DHAN~\cite{Cun_Pun_Shi_2020_DHAN}& 8.94 & 4.80 & 5.67 \\
    AEFNet~\cite{Fu_2021_Auto} & 8.56 & 5.75 & 6.51 \\
    DC-ShadowNet~\cite{jin2021dc} & 8.26 & 3.68 & 4.94 \\
    CANet*~\cite{Chen_2021_CANet} & 7.82 & 5.88 & 5.98 \\
    \midrule[0.5pt]
    CRFormer~(Ours) & \textbf{7.14} & \textbf{3.15} & \textbf{4.25} \\
    	\bottomrule[1.0pt]
    \end{tabular}}}
    \label{tab:srd}
	\vspace{-0.2cm}
\end{table}

\vspace{2pt}
\textbf{On SRD.}
As in~\cite{Cun_Pun_Shi_2020_DHAN,jin2021dc,Fu_2021_Auto, Chen_2021_CANet}, we employ the shadow masks generated by DHAN~\cite{Cun_Pun_Shi_2020_DHAN} for evaluation and compare to current state-of-the-art methods, including DeshadowNet~\cite{qu2017deshadownet}, DSC~\cite{hu2019direction}, DHAN~\cite{Cun_Pun_Shi_2020_DHAN}, AEFNet~\cite{Fu_2021_Auto}, DC-ShadowNet~\cite{jin2021dc}, and CANet~\cite{Chen_2021_CANet}. We report the comparison results in Table~\ref{tab:srd}, which are consistent to those in the ISTD and ASTD datasets.
Our method achieves the best scores of RMSE in shadow regions, non-shadow regions, and the whole image.
Compared to the second best~CANet~\cite{Chen_2021_CANet}, we achieve the 8.7\%, and 23.9\% reduction in RMSE metrics for shadow regions and the entire image, respectively. Some qualitative results are shown in Fig.~\ref{fig:3} (bottom two rows). We observe that our CRFormer has a clear advantage in shadow-removal quality both for the shadow region and the entire image. 

\begin{table}[htbp]\small
    \centering
    \caption{
    Ablation studies on the effectiveness of the key components in our CRFormer on  the AISTD~\cite{Le2019Shadow} testing set.
    }
     \renewcommand\arraystretch{1.1}
    \setlength{\tabcolsep}{4.0mm}{
    {\begin{tabular}{l|c|c|c}
    			\toprule[1.0pt]
    			\multirow{2}{*}{Model}&\multicolumn{1}{c|}{\textbf{S}}& \multicolumn{1}{c}{\textbf{NS}}& \multicolumn{1}{c}{\textbf{All}}\\
    			&RMSE&RMSE&RMSE\\ 
                \midrule[0.5pt]
    			CNN-{ResB} & 6.73 & 2.90 & 3.35\\
    			
    			Single encoder & 6.49& 2.93 & 3.51 \\
                
    			RefineNet only & 7.92 & 2.92 & 3.74\\
    
    			w/o RefineNet & 6.28 & \textbf{2.70} & \textbf{3.29}\\
    			
      		    w/o $L_{ spa}$ & 5.95 & 2.89& 3.39 \\
      		
    			CRFormer (Ours) & \textbf{5.88} & 2.90 & 3.38 \\
    			\bottomrule[1.0pt]
    \end{tabular}}}
    \label{tab:abl_loss}
    \vspace{-.3cm}
\end{table}

\subsection{Ablation Study}
\label{subsec:abl}
In this section, we analyze the impact of each component of our CRFormer in detail where the number of cross-region alignment blocks $N$ is set to $2$. The evaluations are performed on the AISTD~\cite{Le2019Shadow} dataset using different variants.

\vspace{2pt}
\noindent{\textbf{Transformer vs. convolution.} }
To verify the effect of the proposed Transformer layer with region-aware cross-attention, we replace the cross-region alignment blocks of the Transformer layer in CRFormer with convolution-based ResBlocks, resulting in a pure convolutional neural network ``CNN-ResB''.
The quantitative results are shown in Table~\ref{tab:abl_loss}. We observe that CRFormer achieves an RMSE value of 5.88 in the shadow region and outperforms the pure CNN method by 12.6 \%. This study demonstrates the effectiveness of the proposed Transformer layer.

\vspace{2pt}
\noindent
\textbf{Study of architecture and loss.}
We also run experiments to verify the effectiveness of the architectures and loss. It can be seen from the second row of Table~\ref{tab:abl_loss} (``Single encoder''). When the non-shadow path of the dual encoder is removed, \textit{i.e.}, ${F}_{kv} = F_{q}$ in Eq.~(\ref{eq:2}), the value of RMSE in shadow regions increases by 9.4\%, which proves that extra shallow encoder is beneficial to accurately provide non-shadow region features. From rows 3-4, we see that shadow-removal performance drops by 34.7\% in terms of RMSE by only using RefineNet, and our model without RefineNet still achieves competitive results in Table~\ref{tab:aistd}, which also further demonstrates the robust shadow-removal capability of the proposed method. Finally, removing the $L_{spa}$, the performance also slightly drops in the shadow region--RMSE increases by 0.07. These studies verify the effectiveness of the proposed hybrid CNN-Transformer framework.

\begin{table}[htbp]\small
    \centering
    \caption{
    Ablation studies on the effectiveness of the proposed region-aware cross-attention on the AISTD~\cite{Le2019Shadow} testing set.
    }
    \renewcommand\arraystretch{1.1}
    \setlength{\tabcolsep}{4.0mm}{
    {\begin{tabular}{l|c|c|c}
    			\toprule[1.0pt]
    			\multirow{2}{*}{Model}&\multicolumn{1}{c|}{\textbf{S}}& \multicolumn{1}{c}{\textbf{NS}}& \multicolumn{1}{c}{\textbf{All}}\\
    			&RMSE&RMSE&RMSE\\ 
                \midrule[0.5pt]
                Vanilla attention & 6.37 & 2.94 & 3.51\\
    			w/ RCA (S. only) & 6.50 & 2.92 & 3.51\\
    			w/ RCA (Whole) & 6.32 & 2.92 & 3.48\\
    			w/ RCA & \textbf{5.88} & \textbf{2.90} & \textbf{3.38} \\
    			\bottomrule[1.0pt]
    \end{tabular}}}
    \vspace{-.3cm}
    \label{tab:abl_att}
\end{table}
\begin{figure}[htbp]\small
    \centering
	\begin{tabular}{ccc}
		\hspace{-.2cm}\includegraphics[width=0.155\textwidth, height=0.10\textwidth]{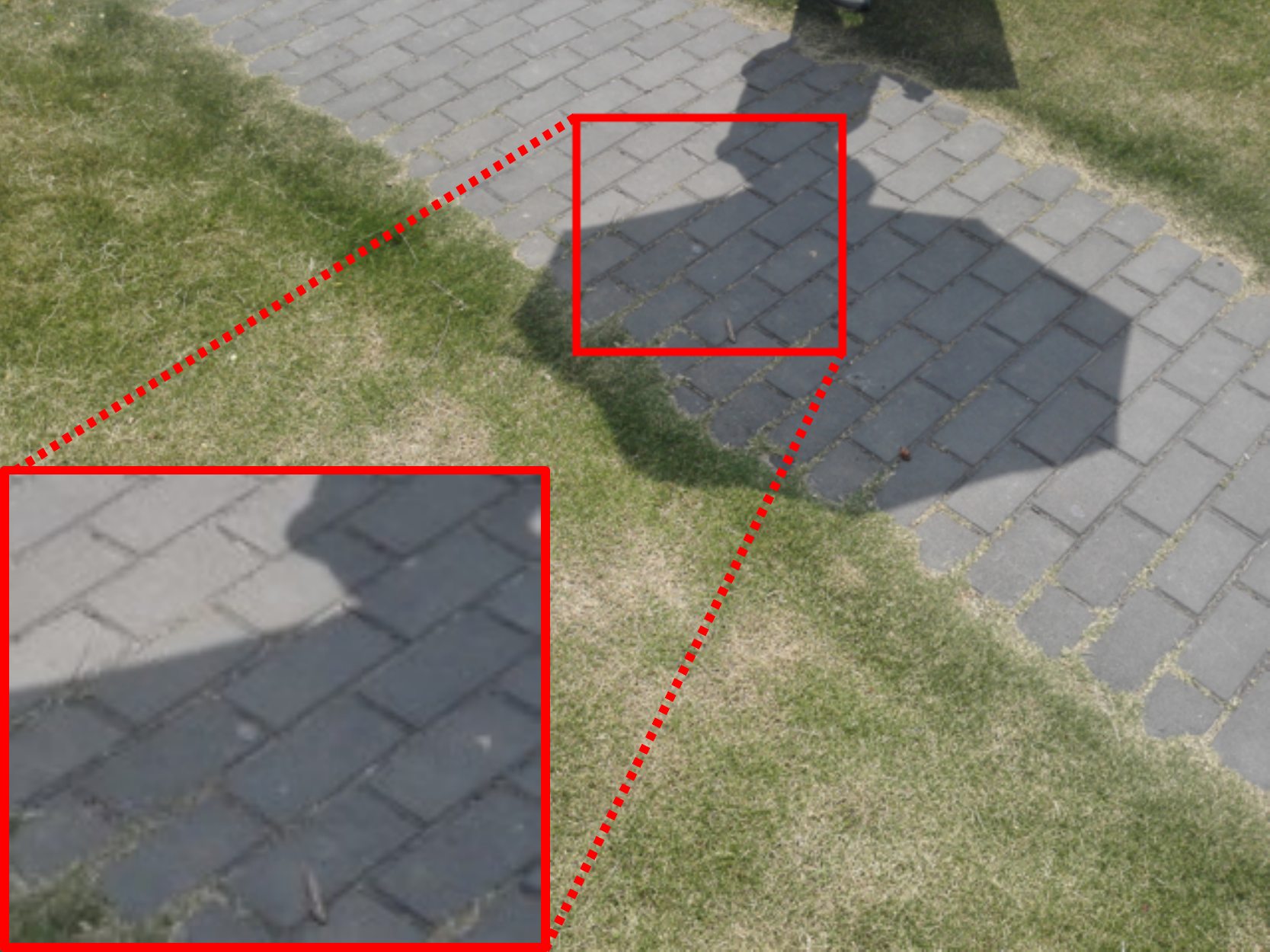} & \hspace{-.45cm}
		\includegraphics[width=0.155\textwidth, height=0.10\textwidth]{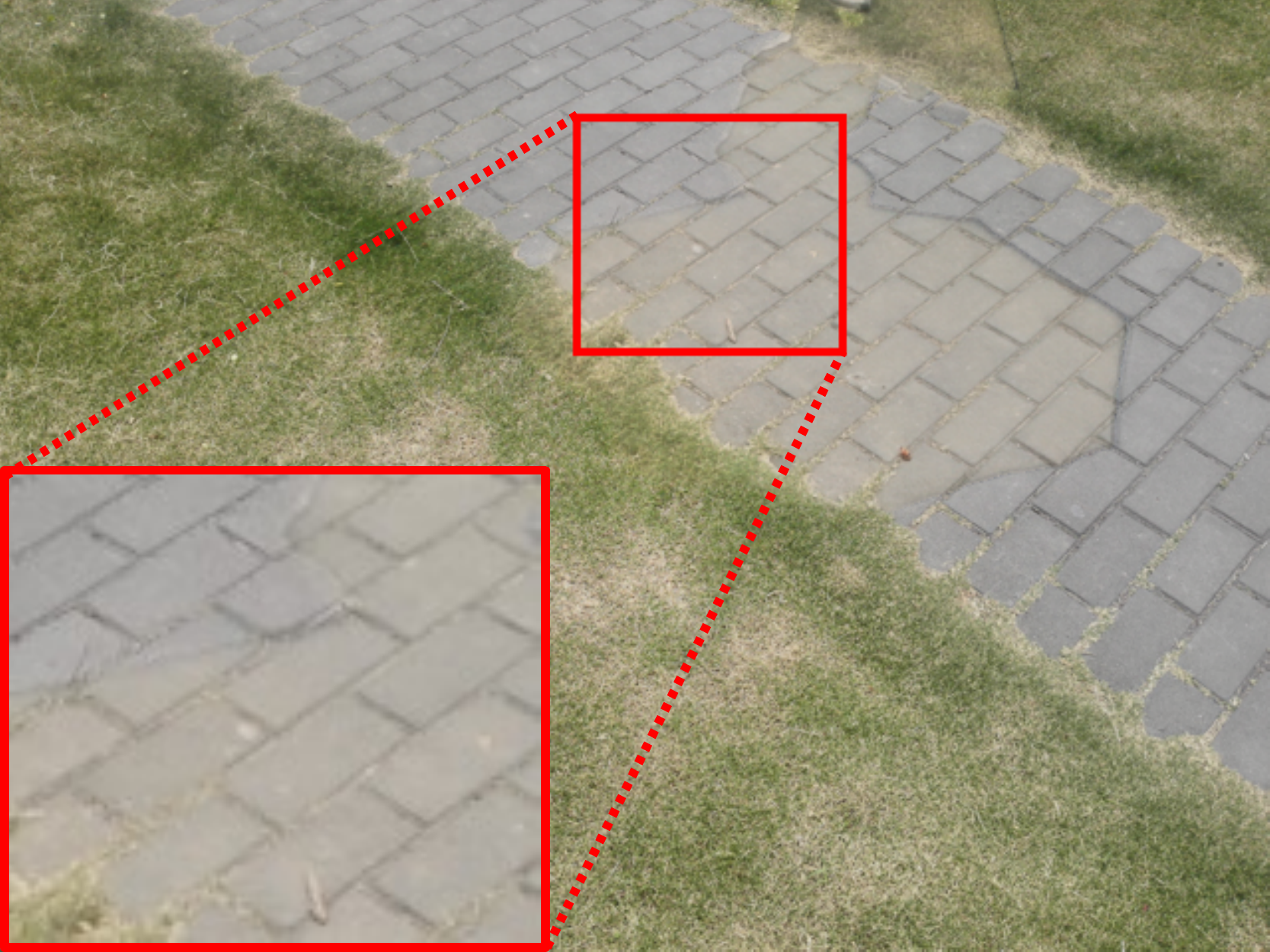}& \hspace{-.45cm}
		\includegraphics[width=0.155\textwidth, height=0.10\textwidth]{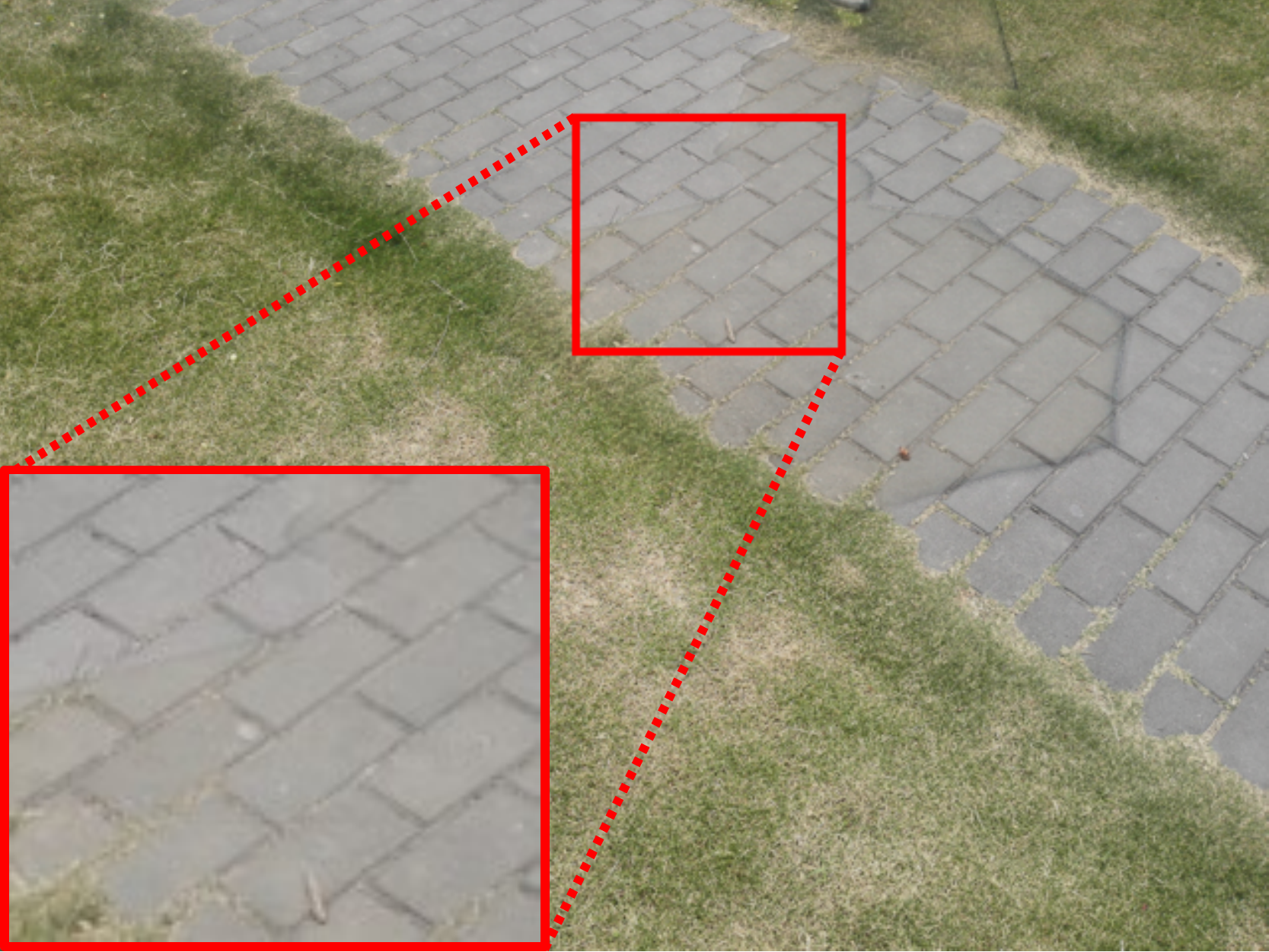}\vspace{-.03cm} \\
		\hspace{-.2cm}\includegraphics[width=0.155\textwidth, height=0.10\textwidth]{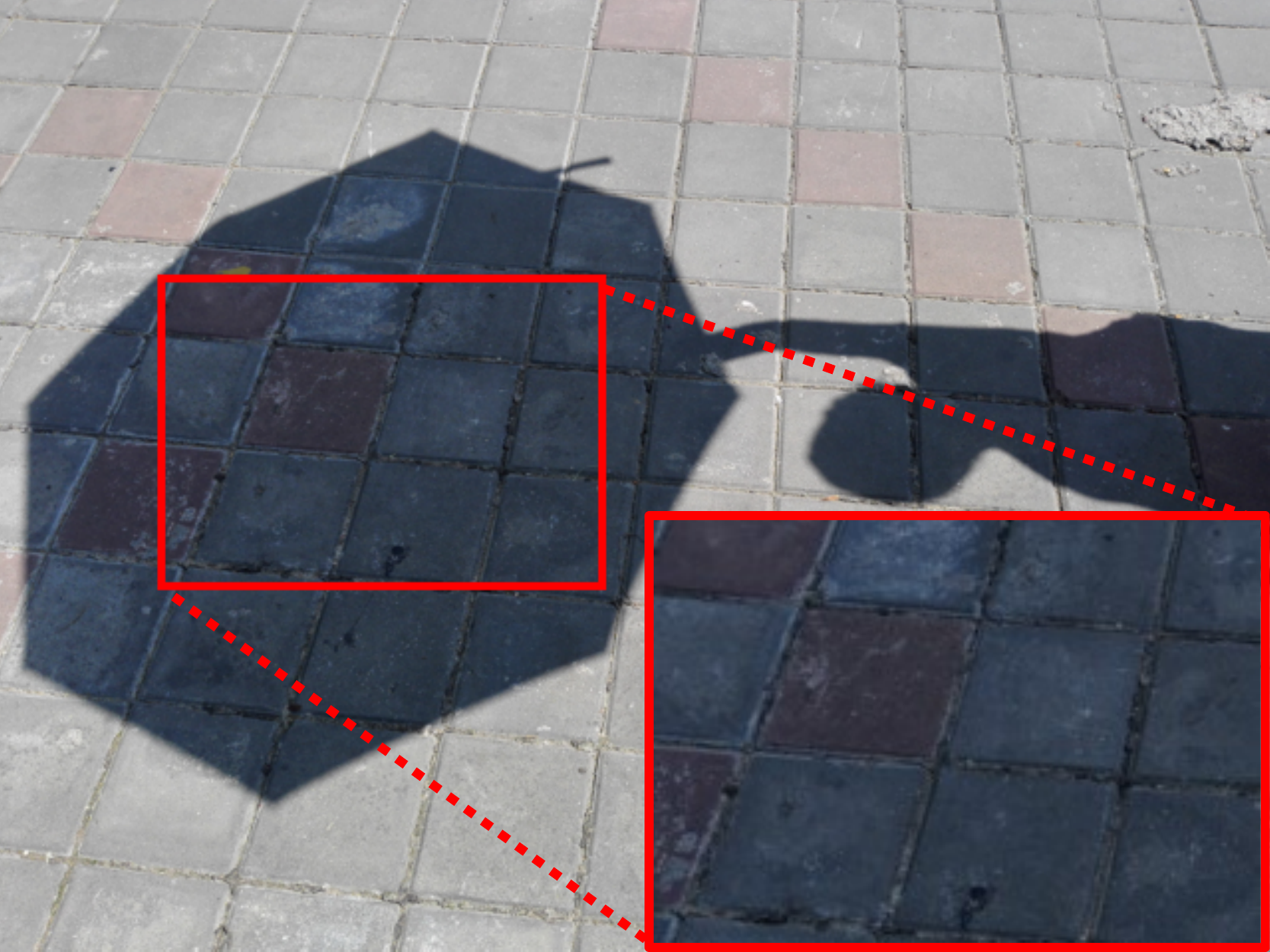} & \hspace{-.45cm}
		\includegraphics[width=0.155\textwidth, height=0.10\textwidth]{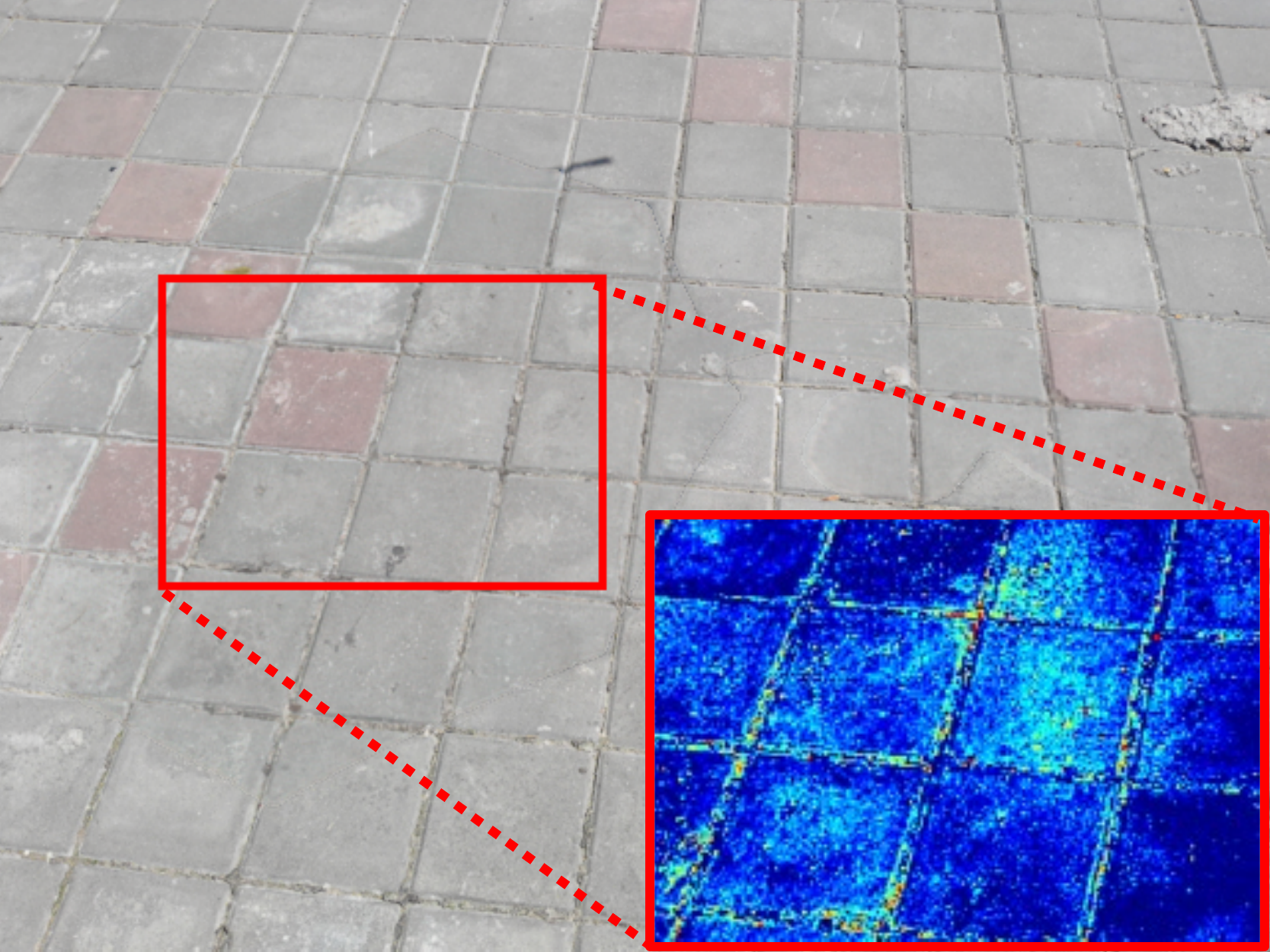}& \hspace{-.45cm}
		\includegraphics[width=0.155\textwidth, height=0.10\textwidth]{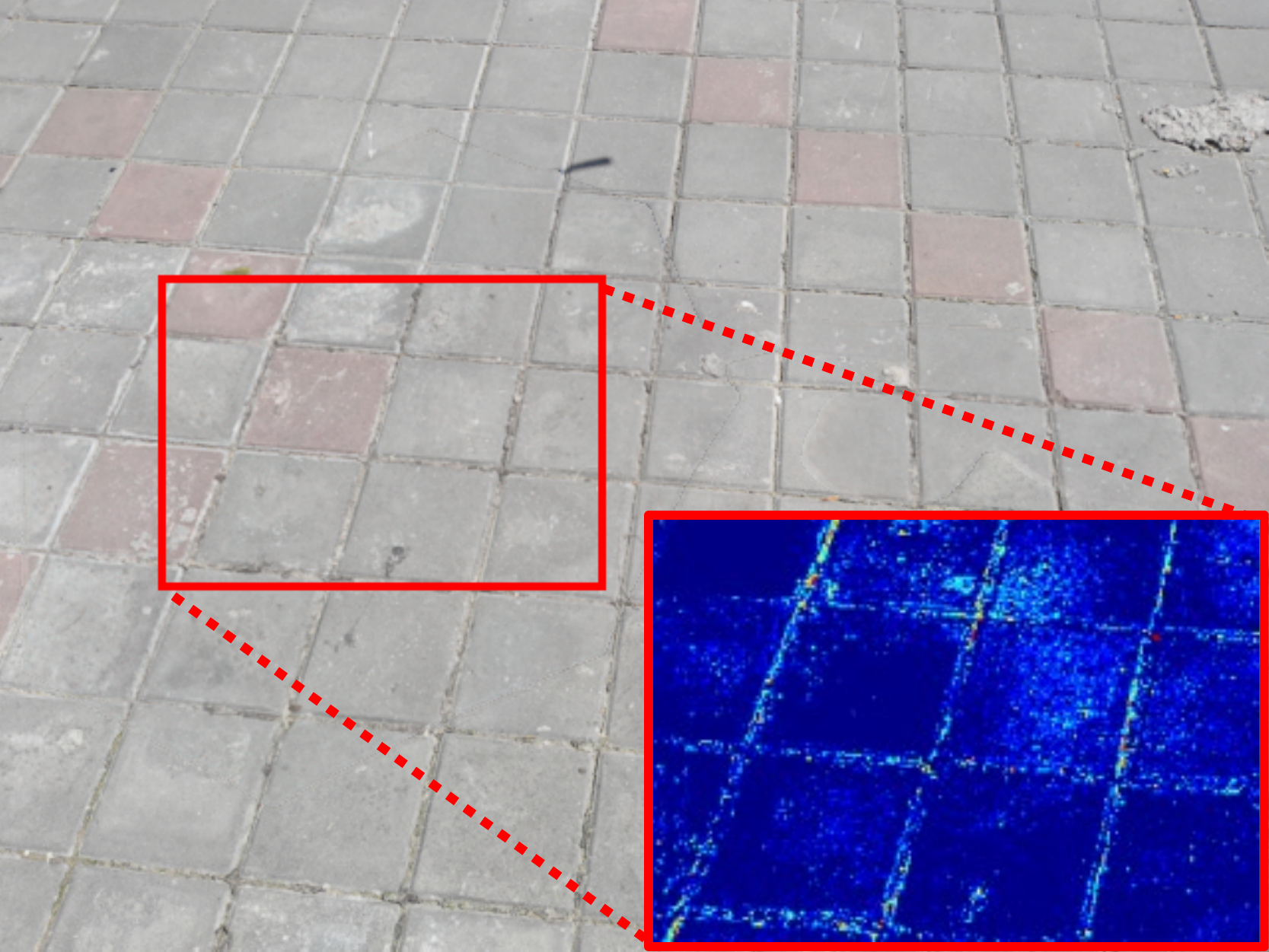}\vspace{-.03cm} \\
	\end{tabular}
	\vspace{-2pt}
    \caption{
    Qualitative comparisons of different attention modules in CRFormer on the AISTD~\cite{Le2019Shadow} dataset. From left to right are the input shadow images, the de-shadowed results of CRFormer using the vanilla attention and the proposed region-aware cross-attention, respectively.
    }
    \label{fig:abl_attention}
	\vspace{-.2cm}
\end{figure}

\begin{figure*}[htp]\small
    \centering
	\begin{tabular}{cccccc}
		\hspace{-.2cm}\includegraphics[width=0.165\textwidth, height=0.100\textwidth]{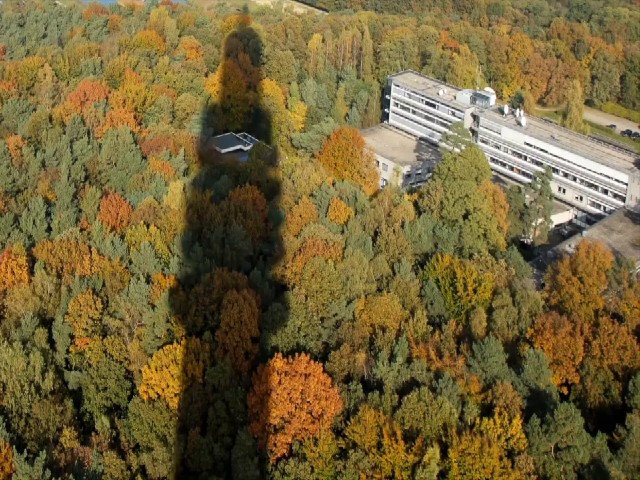} & \hspace{-.45cm}
		\includegraphics[width=0.165\textwidth, height=0.100\textwidth]{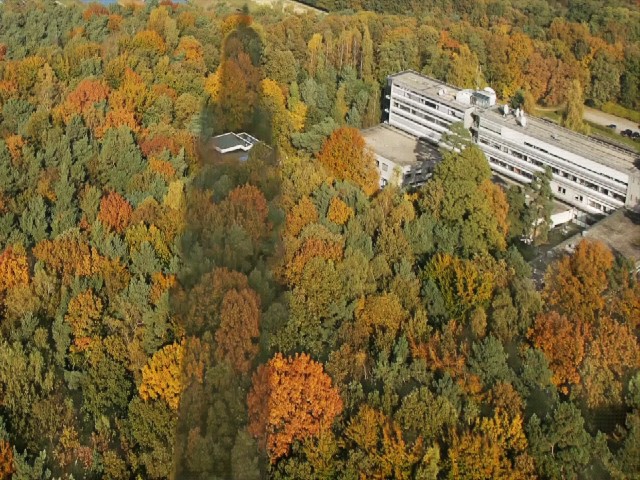}& \hspace{-.45cm}
		\includegraphics[width=0.165\textwidth, height=0.100\textwidth]{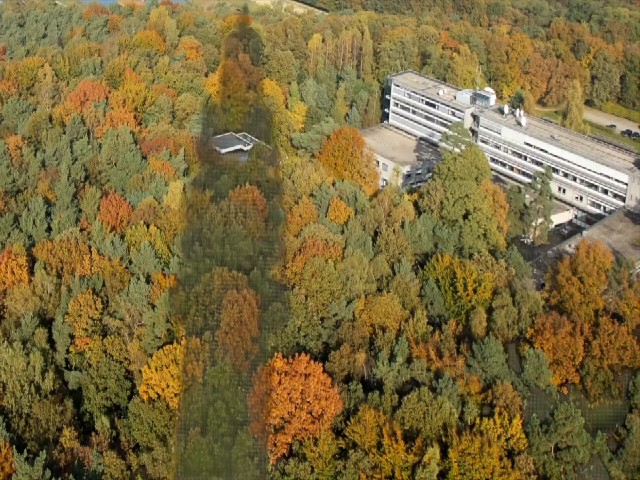}& \hspace{-.45cm}
		\includegraphics[width=0.165\textwidth, height=0.100\textwidth]{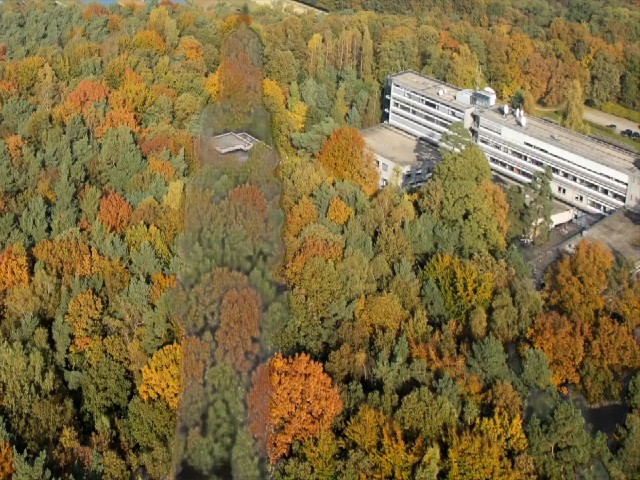}& \hspace{-.45cm}
		\includegraphics[width=0.165\textwidth, height=0.100\textwidth]{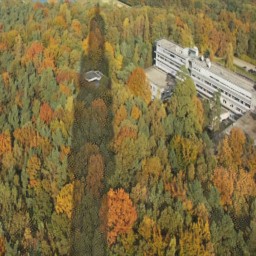}& \hspace{-.45cm}
		\includegraphics[width=0.165\textwidth, height=0.100\textwidth]{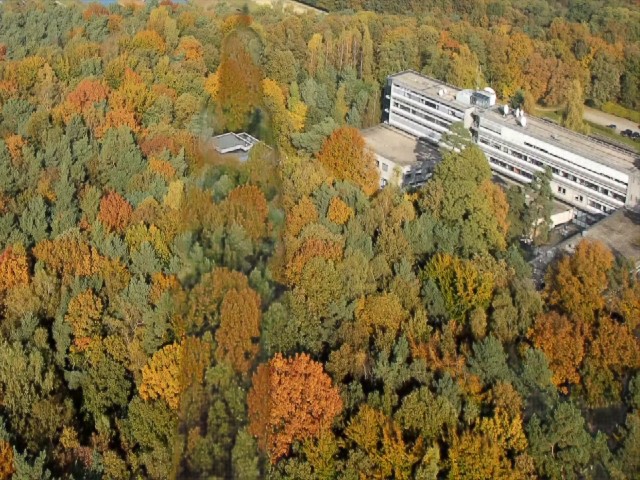}\vspace{-.03cm} \\
		\hspace{-.2cm}\includegraphics[width=0.165\textwidth, height=0.100\textwidth]{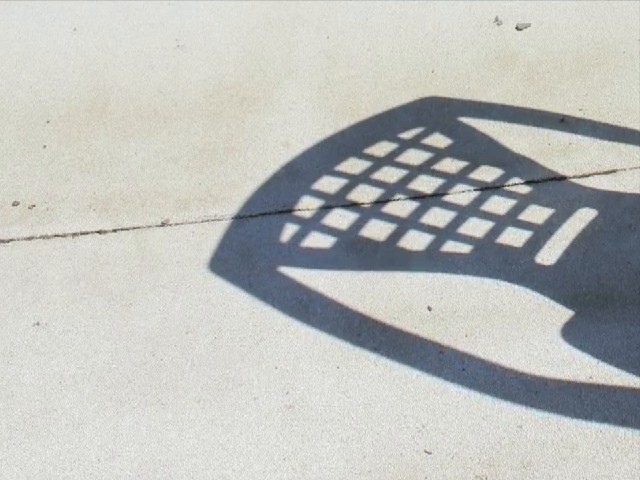} & \hspace{-.45cm}
		\includegraphics[width=0.165\textwidth, height=0.100\textwidth]{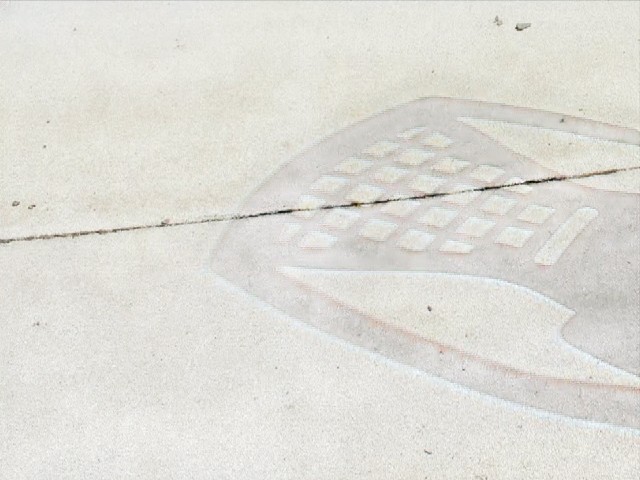}& \hspace{-.45cm}
		\includegraphics[width=0.165\textwidth, height=0.100\textwidth]{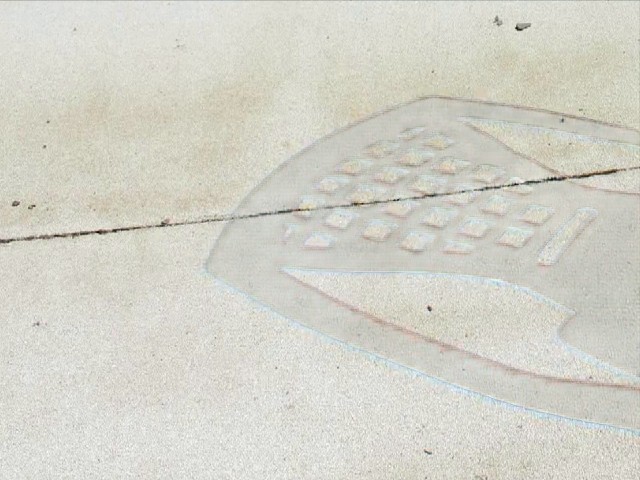}& \hspace{-.45cm}
		\includegraphics[width=0.165\textwidth, height=0.100\textwidth]{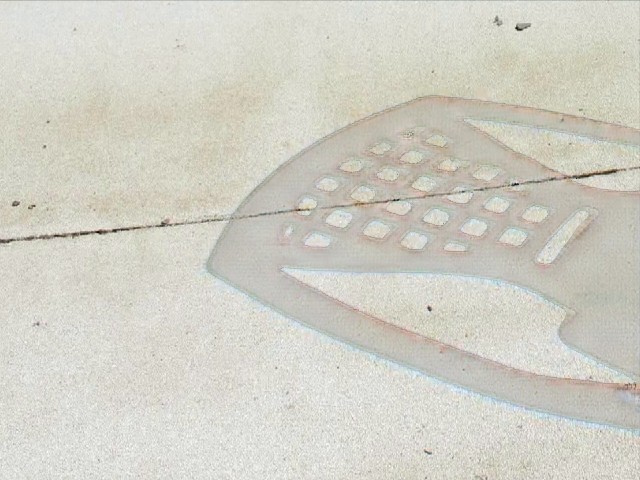}& \hspace{-.45cm}
		\includegraphics[width=0.165\textwidth, height=0.100\textwidth]{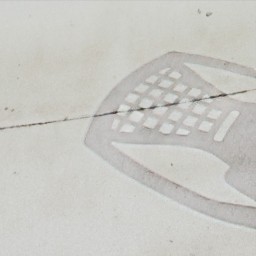}& \hspace{-.45cm}
		\includegraphics[width=0.165\textwidth, height=0.100\textwidth]{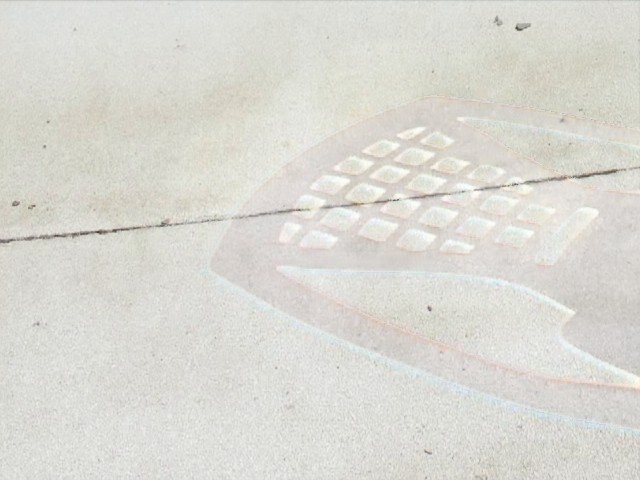}\vspace{-.03cm} \\
		\hspace{-.2cm}  Input & \hspace{-.45cm} {MS-ShadowGAN~\cite{hu2019mask}} & \hspace{-.33cm} {LG-ShadowNet~\cite{liu2021shadow}} & \hspace{-.45cm} {G2R-ShadowNet~\cite{liu2021from}}& \hspace{-.45cm} {DC-ShadowNet~\cite{jin2021dc}} & \hspace{-.45cm} Ours \\  
	\end{tabular}
	\vspace{-4pt}
    \caption{ Qualitative comparisons on two challenging samples (both close~(bottom) and distant~(top) frames) in the Video Shadow Removal~\cite{le2020from} dataset.}
    \label{fig:4}
	\vspace{-.3cm}
\end{figure*}

\noindent
\textbf{Region-aware cross-attention and its variants vs. vanilla attention.}
We further analyze the effect of the proposed region-aware cross-attention (RCA). We first only use the vanilla attention operation in Eq.~(\ref{eq:1}), and the performance drops a lot as shown in Table~\ref{tab:abl_att}. As we discussed above, one reason is that some corrupted features in shadow regions may be irrelevant or even harmful to the de-shadowed features. By using our RCA to tightly integrate the pixel-level features of relevant non-shadow regions into the recovered shadow features, the performance of shadow regions is improved by 7.7\% compared to vanilla attention. Besides, we present the generated shadow-removal results by the vanilla attention and RCA, as shown in Fig.~\ref{fig:abl_attention}. With RCA, the proposed CRFormer can accurately recover the pixel intensity in shadow regions  with less color distortion (red bounding boxes).
Besides, we also try to take the features of different regions, \textit{e.g.}, the shadow region, the whole image, and the non-shadow region into shadow regions. Comparing the last three rows of Table~\ref{tab:abl_att}, we find that the contextual information from the non-shadow region is the best choice to be aggregated into the recovered shadow region features.

\begin{table}[htbp]\small
    \centering
    \caption{
    Ablation studies on the number of cross-region alignment blocks on the AISTD~\cite{Le2019Shadow} testing set. FLOPs are computed by using the 480$\times$640 image as input.
    }
    \renewcommand\arraystretch{1.1}
    \setlength{\tabcolsep}{1.mm}{
    {\begin{tabular}{l|c|c|c|c|c}
    			\toprule[1.0pt]
    			\multirow{2}{*}{Model}& \multirow{2}{*}{\textbf{\#Params.}}& \multirow{2}{*}{\textbf{FLOPs}}& \multicolumn{1}{c}{\textbf{S}}& \multicolumn{1}{c}{\textbf{NS}}& \multicolumn{1}{c}{\textbf{All}}\\
    			&&&RMSE&RMSE&RMSE\\ 
                \midrule[0.5pt]
    	        Transformer-\textit{N1} & 4.76 M & 157.31 G & 6.11 & {2.94} & 3.44\\
    			Transformer-\textit{N2} & 4.89 M & 159.93 G & \textbf{5.88} & \textbf{2.90} & \textbf{3.38}\\
    			Transformer-\textit{N3} & 5.02 M & 162.35 G & 5.93 & 2.94 & 3.43\\
    			\bottomrule[1.0pt]
    \end{tabular}}}
    \label{tab:transformer_N}
    \vspace{-.2cm}
\end{table}

\noindent
\textbf{Study of the number of cross-region alignment blocks.}
The performance of CRFormer is also affected by the number of cross-region alignment blocks (denoted as $N$). By stacking more blocks in the Transformer layer, more times of cross-region alignment will be computed in CRFormer. As shown in Table~\ref{tab:transformer_N}, we found that CRFormer achieves the best performance when using two blocks which is better than using only one block. In addition, using more blocks achieve comparable results as using two blocks and will increase the number of model parameters (0.13M) and computation complexity (2.42G FLOPs).

\begin{table}[htbp]\small
    \centering
    \caption{
    Shadow removal results of our CRFormer compared to state-of-the-art shadow-removal methods on the Video Shadow Removal~\cite{le2020from}. `-' indicates no published results.
    } 
    \renewcommand\arraystretch{1.1}
    \setlength{\tabcolsep}{3.mm}{
    \begin{tabular}{lccc}
    		\toprule[1.0pt]
            Method & RMSE & PSNR & SSIM \\ 
    		\midrule[0.5pt]
            SP+M-Net~\cite{Le2019Shadow} & 22.2 & - & - \\
            Param+M+D-Net~\cite{le2020from} & 20.9 & - & - \\
            Mask-ShadowGAN~\cite{hu2019mask} & 19.6 & 19.47 & 0.850 \\
            LG-ShadowNet~\cite{liu2021shadow} & 18.3 & 19.90 & 0.843 \\
            G2R-ShadowNet~\cite{liu2021from} & 18.8 & 20.00 & 0.838 \\
            DC-ShadowNet~\cite{jin2021dc} & 18.9 & 19.92 & 0.848 \\
    		\midrule[0.5pt]
    		CRFormer (Ours) &\textbf{17.1}& \textbf{21.30} & \textbf{0.854} \\
    		\bottomrule[1.0pt]
    \end{tabular}}
    \label{tab:vsd}
    \vspace{-.2cm}
\end{table}
\subsection{Generalization Ability}
\label{subsec:vsd}
On unseen Video Shadow Removal (VSR)~\cite{le2020from} dataset, we verify the generalization of the proposed method over state-of-the-art algorithms: SP+M-Net~\cite{Le2019Shadow}, Param+M+D-Net~\cite{le2020from}, Mask-ShadowGA-N~\cite{hu2019mask}, LG-ShadowNet~\cite{liu2021shadow}, G2R-ShadowNet~\cite{liu2021from}, and DC-Shadow-Net~\cite{jin2021dc}. All methods are pre-trained on AISTD~\cite{Le2019Shadow} and directly evaluated on the VSR dataset. From the quantitative results in Table~\ref{tab:vsd}, it can be seen that our method significantly outperforms other shadow-removal methods including \textit{supervised method} of SP+M-Net, \textit{weakly-supervised methods} of Param+M+D-Net and G2R-ShadowNet, and \textit{unsupervised methods} of  Mask-ShadowGAN, LG-ShadowNet, and DC-ShadowNet, which proves the strong generalization/practicality of our method. It can also be seen from the qualitative results in Fig.~\ref{fig:4} that by leveraging adequate contextual information from non-shadow regions for restoring the shadowed pixels, our method produces de-shadowed results with less artifacts that are more favorable and realistic than other methods.

\section{Conclusion}
In this paper, we proposed CRFormer, a novel hybrid CNN-transformer framework, for high-quality shadow removal. It is able to capture adequate contextual information from non-shadow regions to help restore the shadow-free pixel intensity of shadow regions owing to the newly designed region-aware cross-attention operation.
Different from existing attention computation in transformers, the region-aware cross-attention is a one-way attention computed only from the non-shadow region to the shadow region for de-shadowed feature aggregation.
Experimental results show that our CRFormer produces superior results compared to the state-of-the-arts on ISTD, AISTD, SRD, and Video Shadow Removal datasets.



{\small
	\bibliographystyle{ieee_fullname}
	\bibliography{sample-base}

\begin{thebibliography}{10}\itemsep=-1pt

\bibitem{Bako16documentS}
Steve Bako, Soheil Darabi, Eli Shechtman, Jue Wang, Kalyan Sunkavalli, and
  Pradeep Sen.
\newblock Removing shadows from images of documents.
\newblock pages 173--183, 2016.

\bibitem{botach2021end}
Adam Botach, Evgenii Zheltonozhskii, and Chaim Baskin.
\newblock End-to-end referring video object segmentation with multimodal
  transformers.
\newblock {\em arXiv preprint arXiv:2111.14821}, 2021.

\bibitem{Caelles2017OneShotVO}
Sergi Caelles, Kevis-Kokitsi Maninis, Jordi Pont-Tuset, Laura Leal-Taix{\'e},
  Daniel Cremers, and Luc~Van Gool.
\newblock One-shot video object segmentation.
\newblock pages 5320--5329, 2017.

\bibitem{carion2020end}
Nicolas Carion, Francisco Massa, Gabriel Synnaeve, Nicolas Usunier, Alexander
  Kirillov, and Sergey Zagoruyko.
\newblock End-to-end object detection with transformers.
\newblock In {\em Eur. Conf. Comput. Vis.}, pages 213--229. Springer, 2020.

\bibitem{Chen_2021_CANet}
Zipei Chen, Chengjiang Long, Ling Zhang, and Chunxia Xiao.
\newblock Canet: A context-aware network for shadow removal.
\newblock In {\em Int. Conf. Comput. Vis.}, pages 4743--4752, October 2021.

\bibitem{Cun_Pun_Shi_2020_DHAN}
Xiaodong Cun, Chi-Man Pun, and Cheng Shi.
\newblock Towards ghost-free shadow removal via dual hierarchical aggregation
  network and shadow matting gan.
\newblock volume~34, pages 10680--10687, Apr. 2020.

\bibitem{dai2021up}
Zhigang Dai, Bolun Cai, Yugeng Lin, and Junying Chen.
\newblock Up-detr: Unsupervised pre-training for object detection with
  transformers.
\newblock In {\em IEEE Conf. Comput. Vis. Pattern Recog.}, pages 1601--1610,
  2021.

\bibitem{Devlin2019BERTPO}
Jacob Devlin, Ming-Wei Chang, Kenton Lee, and Kristina Toutanova.
\newblock Bert: Pre-training of deep bidirectional transformers for language
  understanding.
\newblock In {\em NAACL}, pages 4171--4186, 2019.

\bibitem{dosovitskiy2021an}
Alexey Dosovitskiy, Lucas Beyer, Alexander Kolesnikov, Dirk Weissenborn,
  Xiaohua Zhai, Thomas Unterthiner, Mostafa Dehghani, Matthias Minderer, Georg
  Heigold, Sylvain Gelly, Jakob Uszkoreit, and Neil Houlsby.
\newblock An image is worth 16x16 words: Transformers for image recognition at
  scale.
\newblock In {\em Int. Conf. Learn. Represent.}, 2021.

\bibitem{Fu_2021_Auto}
Lan Fu, Changqing Zhou, Qing Guo, Felix Juefei-Xu, Hongkai Yu, Wei Feng, Yang
  Liu, and Song Wang.
\newblock Auto-exposure fusion for single-image shadow removal.
\newblock In {\em IEEE Conf. Comput. Vis. Pattern Recog.}, pages 10571--10580,
  June 2021.

\bibitem{Girshick_2014_rcnn}
Ross Girshick, Jeff Donahue, Trevor Darrell, and Jitendra Malik.
\newblock Rich feature hierarchies for accurate object detection and semantic
  segmentation.
\newblock In {\em IEEE Conf. Comput. Vis. Pattern Recog.}, pages 580--587,
  2014.

\bibitem{Girshick2015FaRcnn}
Ross~B. Girshick.
\newblock Fast r\-cnn.
\newblock pages 1440--1448, 2015.

\bibitem{gong2014interactive}
Han Gong and Darren Cosker.
\newblock Interactive shadow removal and ground truth for variable scene
  categories.
\newblock In {\em Brit. Mach. Vis. Conf.}, pages 1--11, 2014.

\bibitem{gryka2015learning}
Maciej Gryka, Michael Terry, and Gabriel~J Brostow.
\newblock Learning to remove soft shadows.
\newblock {\em ACM Trans. Graph.}, 34(5):153, 2015.

\bibitem{guo2020zero}
Chunle Guo, Chongyi Li, Jichang Guo, Chen~Change Loy, Junhui Hou, Sam Kwong,
  and Runmin Cong.
\newblock Zero-reference deep curve estimation for low-light image enhancement.
\newblock In {\em IEEE Conf. Comput. Vis. Pattern Recog.}, pages 1780--1789,
  2020.

\bibitem{guo2012paired}
Ruiqi Guo, Qieyun Dai, and Derek Hoiem.
\newblock Paired regions for shadow detection and removal.
\newblock {\em IEEE Trans. Pattern Anal. Mach. Intell.}, 35(12):2956--2967,
  2012.

\bibitem{guo2021image}
Zonghui Guo, Dongsheng Guo, Haiyong Zheng, Zhaorui Gu, Bing Zheng, and Junyu
  Dong.
\newblock Image harmonization with transformer.
\newblock In {\em Int. Conf. Comput. Vis.}, pages 14870--14879, 2021.

\bibitem{He_21mmshadowFace}
Yingqing He, Yazhou Xing, Tianjia Zhang, and Qifeng Chen.
\newblock Unsupervised portrait shadow removal via generative priors.
\newblock In {\em ACM Int. Conf. Multimedia}, page 236–244, 2021.

\bibitem{hu2019direction}
Xiaowei Hu, Chi-Wing Fu, Lei Zhu, Jing Qin, and Pheng-Ann Heng.
\newblock Direction-aware spatial context features for shadow detection and
  removal.
\newblock {\em IEEE Trans. Pattern Anal. Mach. Intell.}, 42(11):2795--2808,
  2019.

\bibitem{hu2019mask}
Xiaowei Hu, Yitong Jiang, Chi-Wing Fu, and Pheng-Ann Heng.
\newblock Mask-shadowgan: Learning to remove shadows from unpaired data.
\newblock In {\em Int. Conf. Comput. Vis.}, pages 2472--2481, 2019.

\bibitem{jin2021dc}
Yeying Jin, Aashish Sharma, and Robby~T Tan.
\newblock Dc-shadownet: Single-image hard and soft shadow removal using
  unsupervised domain-classifier guided network.
\newblock In {\em Int. Conf. Comput. Vis.}, pages 5027--5036, 2021.

\bibitem{kim2021hotr}
Bumsoo Kim, Junhyun Lee, Jaewoo Kang, Eun-Sol Kim, and Hyunwoo~J Kim.
\newblock Hotr: End-to-end human-object interaction detection with
  transformers.
\newblock In {\em IEEE Conf. Comput. Vis. Pattern Recog.}, pages 74--83, 2021.

\bibitem{Kingma2015AdamAM}
Diederik~P. Kingma and Jimmy Ba.
\newblock Adam: A method for stochastic optimization.
\newblock {\em arXiv preprint arXiv:1412.6980}, 2015.

\bibitem{le2019weakly}
Hieu Le, Bento Goncalves, Dimitris Samaras, and Heather Lynch.
\newblock Weakly labeling the antarctic: The penguin colony case.
\newblock In {\em IEEE Conf. Comput. Vis. Pattern Recog. Worksh.}, pages
  18--25, 2019.

\bibitem{Le2019Shadow}
Hieu Le and Dimitris Samaras.
\newblock Shadow removal via shadow image decomposition.
\newblock In {\em Int. Conf. Comput. Vis.}, pages 8578--8587, 2019.

\bibitem{le2020from}
Hieu Le and Dimitris Samaras.
\newblock From shadow segmentation to shadow removal.
\newblock In {\em Eur. Conf. Comput. Vis.}, pages 264--281, 2020.

\bibitem{Le_tpami21}
Hieu Le and Dimitris Samaras.
\newblock Physics-based shadow image decomposition for shadow removal.
\newblock {\em IEEE Trans. Pattern Anal. Mach. Intell.}, 2021.

\bibitem{li2021pose}
Ke Li, Shijie Wang, Xiang Zhang, Yifan Xu, Weijian Xu, and Zhuowen Tu.
\newblock Pose recognition with cascade transformers.
\newblock In {\em IEEE Conf. Comput. Vis. Pattern Recog.}, pages 1944--1953,
  2021.

\bibitem{li2021diverse}
Yulin Li, Jianfeng He, Tianzhu Zhang, Xiang Liu, Yongdong Zhang, and Feng Wu.
\newblock Diverse part discovery: Occluded person re-identification with
  part-aware transformer.
\newblock In {\em IEEE Conf. Comput. Vis. Pattern Recog.}, pages 2898--2907,
  2021.

\bibitem{li2021revisiting}
Zhaoshuo Li, Xingtong Liu, Nathan Drenkow, Andy Ding, Francis~X Creighton,
  Russell~H Taylor, and Mathias Unberath.
\newblock Revisiting stereo depth estimation from a sequence-to-sequence
  perspective with transformers.
\newblock In {\em IEEE Conf. Comput. Vis. Pattern Recog.}, pages 6197--6206,
  2021.

\bibitem{Li2018LearningII}
Zhengqi Li and Noah Snavely.
\newblock Learning intrinsic image decomposition from watching the world.
\newblock pages 9039--9048, 2018.

\bibitem{Lin2020documentS}
Yun-Hsuan Lin, Wen-Chin Chen, and Yung-Yu Chuang.
\newblock Bedsr-net: A deep shadow removal network from a single document
  image.
\newblock In {\em IEEE Conf. Comput. Vis. Pattern Recog.}, pages 12902--12911,
  2020.

\bibitem{Liu2020UnsupervisedLF}
Yunfei Liu, Yu Li, Shaodi You, and Feng Lu.
\newblock Unsupervised learning for intrinsic image decomposition from a single
  image.
\newblock pages 3245--3254, 2020.

\bibitem{liuswinT}
Ze Liu, Yutong Lin, Yue Cao, Han Hu, Yixuan Wei, Zheng Zhang, Stephen Lin, and
  Baining Guo.
\newblock Swin transformer: Hierarchical vision transformer using shifted
  windows.
\newblock In {\em IEEE Conf. Comput. Vis. Pattern Recog.}, pages 10012--10022,
  2021.

\bibitem{liu2021shadow}
Zhihao Liu, Hui Yin, Yang Mi, Mengyang Pu, and Song Wang.
\newblock Shadow removal by a lightness-guided network with training on
  unpaired data.
\newblock {\em IEEE Trans. Image Process.}, 30:1853--1865, 2021.

\bibitem{liu2021from}
Zhihao Liu, Hui Yin, Xinyi Wu, Zhenyao Wu, Yang Mi, and Song Wang.
\newblock From shadow generation to shadow removal.
\newblock In {\em IEEE Conf. Comput. Vis. Pattern Recog.}, pages 4927--4936,
  2021.

\bibitem{pu2022edter}
Mengyang Pu, Yaping Huang, Yuming Liu, Qingji Guan, and Haibin Ling.
\newblock Edter: Edge detection with transformer.
\newblock {\em arXiv preprint arXiv:2203.08566}, 2022.

\bibitem{qu2017deshadownet}
Liangqiong Qu, Jiandong Tian, Shengfeng He, Yandong Tang, and Rynson~WH Lau.
\newblock Deshadownet: A multi-context embedding deep network for shadow
  removal.
\newblock In {\em IEEE Conf. Comput. Vis. Pattern Recog.}, pages 4067--4075,
  2017.

\bibitem{Shechtman2016Appearance}
Shechtman, Eli, Li-Qian, Sunkavalli, Kalyan, Shi-Min, Wang, and Jue.
\newblock Appearance harmonization for single image shadow removal.
\newblock volume~35, pages 189--197, 2016.

\bibitem{Shelhamer2017FCN}
Evan Shelhamer, Jonathan Long, and Trevor Darrell.
\newblock Fully convolutional networks for semantic segmentation.
\newblock {\em IEEE Trans. Pattern Anal. Mach. Intell.}, 39:640--651, 2017.

\bibitem{shor2008shadow}
Yael Shor and Dani Lischinski.
\newblock The shadow meets the mask: Pyramid‐based shadow removal.
\newblock {\em Comput. Graph. Forum}, 27:577--586, 2008.

\bibitem{vaswani2017attention}
Ashish Vaswani, Noam Shazeer, Niki Parmar, Jakob Uszkoreit, Llion Jones,
  Aidan~N Gomez, {\L}ukasz Kaiser, and Illia Polosukhin.
\newblock Attention is all you need.
\newblock volume~30, 2017.

\bibitem{vicente2016large}
Tom{\'a}s F~Yago Vicente, Le Hou, Chen-Ping Yu, Minh Hoai, and Dimitris
  Samaras.
\newblock Large-scale training of shadow detectors with noisily-annotated
  shadow examples.
\newblock In {\em Eur. Conf. Comput. Vis.}, pages 816--832. Springer, 2016.

\bibitem{wang2018stacked}
Jifeng Wang, Xiang Li, and Jian Yang.
\newblock Stacked conditional generative adversarial networks for jointly
  learning shadow detection and shadow removal.
\newblock In {\em IEEE Conf. Comput. Vis. Pattern Recog.}, pages 1788--1797,
  2018.

\bibitem{Wu_2021_CVPR}
Xinyi Wu, Zhenyao Wu, Hao Guo, Lili Ju, and Song Wang.
\newblock Dannet: A one-stage domain adaptation network for unsupervised
  nighttime semantic segmentation.
\newblock In {\em IEEE Conf. Comput. Vis. Pattern Recog.}, pages 15769--15778,
  June 2021.

\bibitem{zhang2021few}
Gengwei Zhang, Guoliang Kang, Yi Yang, and Yunchao Wei.
\newblock Few-shot segmentation via cycle-consistent transformer.
\newblock volume~34, 2021.

\bibitem{zhang2015shadow}
Ling Zhang, Qing Zhang, and Chunxia Xiao.
\newblock Shadow remover: Image shadow removal based on illumination recovering
  optimization.
\newblock {\em IEEE Trans. Image Process.}, 24(11):4623--4636, 2015.

\bibitem{Zhang2018LPIPS}
Richard Zhang, Phillip Isola, Alexei~A. Efros, Eli Shechtman, and Oliver Wang.
\newblock The unreasonable effectiveness of deep features as a perceptual
  metric.
\newblock pages 586--595, 2018.

\bibitem{zhang2019shadowgan}
Shuyang Zhang, Runze Liang, and Miao Wang.
\newblock Shadowgan: Shadow synthesis for virtual objects with conditional
  adversarial networks.
\newblock {\em Computational Visual Media}, 5(1):105--115, 2019.

\bibitem{zheng2021rethinking}
Sixiao Zheng, Jiachen Lu, Hengshuang Zhao, Xiatian Zhu, Zekun Luo, Yabiao Wang,
  Yanwei Fu, Jianfeng Feng, Tao Xiang, Philip~HS Torr, et~al.
\newblock Rethinking semantic segmentation from a sequence-to-sequence
  perspective with transformers.
\newblock In {\em IEEE Conf. Comput. Vis. Pattern Recog.}, pages 6881--6890,
  2021.

\bibitem{zhu2018bidirectional}
Lei Zhu, Zijun Deng, Xiaowei Hu, Chi-Wing Fu, Xuemiao Xu, Jing Qin, and
  Pheng-Ann Heng.
\newblock Bidirectional feature pyramid network with recurrent attention
  residual modules for shadow detection.
\newblock In {\em Eur. Conf. Comput. Vis.}, pages 121--136, 2018.

\bibitem{zou2021end}
Cheng Zou, Bohan Wang, Yue Hu, Junqi Liu, Qian Wu, Yu Zhao, Boxun Li, Chenguang
  Zhang, Chi Zhang, Yichen Wei, et~al.
\newblock End-to-end human object interaction detection with hoi transformer.
\newblock In {\em IEEE Conf. Comput. Vis. Pattern Recog.}, pages 11825--11834,
  2021.

\end{thebibliography}
}
	
\end{document}